\documentclass[10pt,twocolumn,letterpaper]{article}

\usepackage[pagenumbers]{cvpr}
\usepackage{times}
\usepackage{epsfig}
\usepackage{graphicx}
\usepackage{amsmath}
\usepackage{amssymb}
\usepackage{multirow}
\usepackage{dsfont}
\usepackage[normalem]{ulem}
\useunder{\uline}{\ul}{}
\usepackage{booktabs}
\usepackage{rotating}
\usepackage{array}
\usepackage{tabularray}
\usepackage{makecell}
\usepackage{cellspace}
\usepackage{wrapfig}
\usepackage{xr-hyper}
\usepackage{subcaption,dcolumn}

\usepackage[skip=0pt,font=small]{caption}

\DeclareMathOperator*{\argmax}{arg\,max}
\newcommand{\squishlist}{
	\begin{list}{$\bullet$}
		{ \setlength{\itemsep}{0pt}
			\setlength{\parsep}{1pt}
			\setlength{\topsep}{1pt}
			\setlength{\partopsep}{0pt}
			\setlength{\leftmargin}{1.5em}
			\setlength{\labelwidth}{1em}
			\setlength{\labelsep}{0.5em} } }
\newcommand{\squishend}{\end{list} 
}
\newcommand{\specialfootnotetext}[1]{
  \begingroup
  \def\thefootnote{\fnsymbol{footnote}}
  \footnotetext[0]{#1}
  \endgroup
}

\newcolumntype{d}[1]{D..{#1}}
\usepackage[table,dvipsnames]{xcolor}
\usepackage{comment}

\definecolor{cvprblue}{rgb}{0.21,0.49,0.74}
\usepackage[pagebackref,breaklinks,colorlinks,citecolor=cvprblue]{hyperref}
\usepackage{float}

\def\paperID{16798} % *** Enter the Paper ID here
\def\confName{CVPR}
\def\confYear{2024}

\begin{document}

%%%%%%%%% TITLE
\title{SnAG: \underline{S}calable a\underline{n}d \underline{A}ccurate Video \underline{G}rounding}
\author{Fangzhou Mu\textsuperscript{*}, Sicheng Mo\textsuperscript{*}, Yin Li \\[0.05in] University of Wisconsin-Madison}

\maketitle

\specialfootnotetext{* indicates equal contribution}

%%%%%%%%% ABSTRACT
\begin{abstract}
Temporal grounding of text descriptions in videos is a central problem in vision-language learning and video understanding. Existing methods often prioritize accuracy over scalability --- they have been optimized for grounding only a few text queries within short videos, and fail to scale up to long videos with hundreds of queries. In this paper, we study the effect of cross-modal fusion on the scalability of video grounding models. Our analysis establishes late fusion as a more cost-effective fusion scheme for long-form videos with many text queries. Moreover, it leads us to a novel, video-centric sampling scheme for efficient training. Based on these findings, we present SnAG, a simple baseline for scalable and accurate video grounding. Without bells and whistles, SnAG is 43\% more accurate and 1.5$\times$ faster than CONE, a state of the art for long-form video grounding on the challenging MAD dataset, while achieving highly competitive results on short videos. Our code is available at~\url{https://github.com/fmu2/snag_release}.

\end{abstract}

%%%%%%%%% BODY TEXT
% intro
\section{Introduction}

Localizing moments within an untrimmed video based on text descriptions, also known as temporal video grounding, involves joint reasoning about visual, textual and temporal information, and thus remains a challenging problem in video understanding. Video grounding is gaining traction amid the rapid advances in vision-language learning, with a range of applications such as intelligent personal assistants, human-robot interaction, and interactive video editing.

Despite recent progress, one largely overlooked aspect of video grounding is the \emph{scalability} of models with respect to video duration and the number of text descriptions (\ie, queries). Previous methods (\eg,~\cite{zhang20202dtan,soldan2021vlgnet}) are primarily designed for grounding only a few queries within short video snippets. However, the growing availability of long videos (\eg, on streaming platforms) and demand to query their rich contents introduce a paradigm shift, necessitating efficient grounding of large volumes of queries in long videos. On latest long-form video grounding benchmarks~\cite{grauman2022ego4d,soldan2022mad}, these methods struggle to scale up due to their high computational cost, and suffer major performance degradation.

\begin{figure}
\centering
\includegraphics[width=0.7\linewidth]{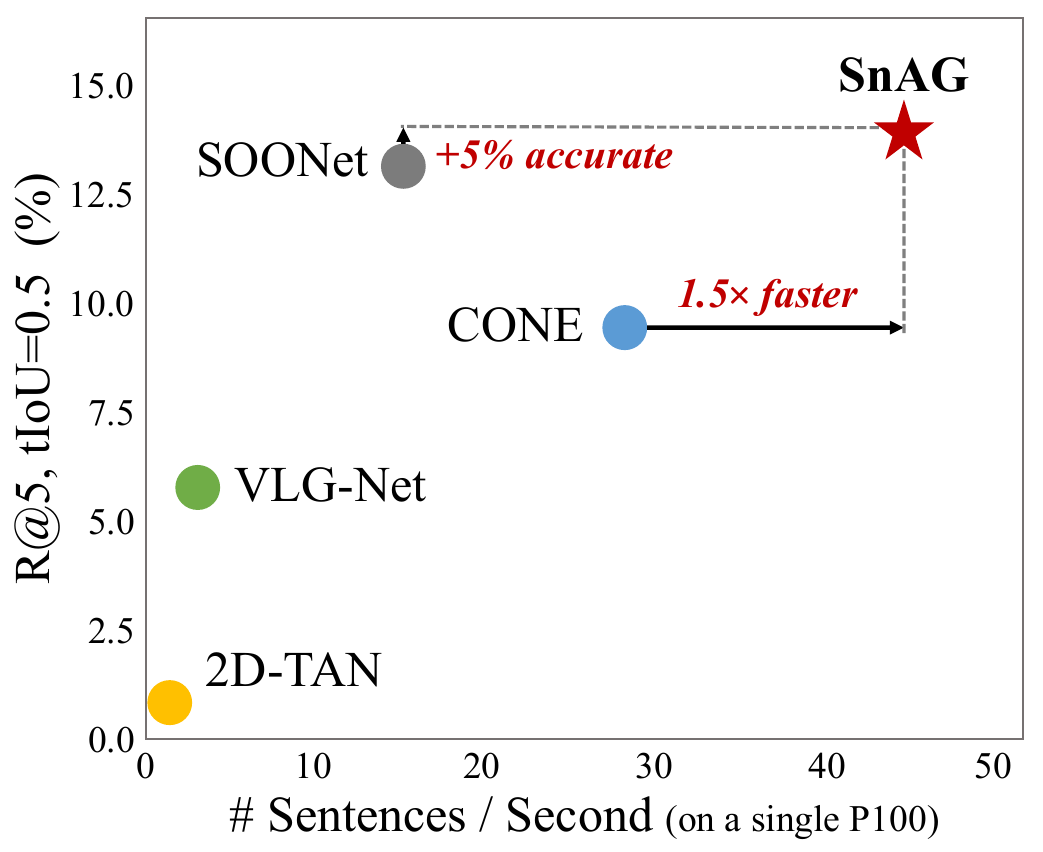}
\vspace{0.5em}
\caption{SnAG achieves the best accuracy and throughput simultaneously on the MAD dataset~\cite{soldan2022mad} for long-form video grounding.}
\vspace{-1.5em}
\label{fig:teaser}
\end{figure}

In this paper, we take a principled approach to studying the problem of scalable video grounding, focusing on the design of cross-modal fusion---a key model component that combines visual and textual information. Specifically, we present an analysis that connects the computational cost of video grounding models to their fusion schemes. Our analysis and empirical results challenge the widely adopted early fusion, showing that late fusion scales better in inference by amortizing the cost of long video processing among text queries. Importantly, this insight further motivates the design of a novel, video-centric sampling scheme for scalable training, without compromising grounding accuracy.

Following these findings, we present SnAG, a scalable and accurate model for long-form video grounding. SnAG features a minimalist, late-fusion design for scalable inference, while supporting video-centric sampling for scalable training. Notwithstanding its simplicity, SnAG achieves superior results across all benchmarks designed for long-form video grounding. Without bells and whistles, SnAG attains 13.75\% R@5, tIoU=0.5 on MAD~\cite{soldan2022mad} with an inference speed of 45.3 queries per second on a P100 GPU (Fig.~\ref{fig:teaser}), 43\% more accurate and 1.5$\times$ faster than the concurrent method of CONE~\cite{hou2022cone}. In the meantime, SnAG is highly competitive on short-video benchmarks. It achieves 46.26\% R@1, tIoU=0.7 on Charades-STA~\cite{sigurdsson2016charades}, outperforming the previous state of the art (SSRN~\cite{zhu2023ssrn}) by 3.61 absolute percentage points while being simpler in model design.

\smallskip \noindent \textbf{Contributions.} \textbf{(1)} We present one of the first studies to understand the effect of fusion schemes on the computational cost of video grounding. \textbf{(2)} Our analysis offers new insights and leads to a new training scheme for scalable video grounding. \textbf{(3)} Based on our findings, we further deliver a simple model that attains superior results across major benchmarks for long-form video grounding while achieving significant efficiency gains in both training and inference.

% related work
\section{Related Work}

\noindent \textbf{Video grounding}. Previous works on video grounding can be categorized based on whether proposals are generated as moment candidates (two-stage \vs.\ single-stage), and how text queries inform action localization (cross-modal fusion).

\textit{Two-stage} methods first generate temporal segments as proposals, then score their ``actionness" and optionally refine their boundaries. Early works~\cite{anne2017mcn,gao2017ctrl} densely sample proposals using sliding windows and score them independently. One line of work~\cite{xu2019qspn,chen2019sap,xiao2021bpnet,liu2021apgn, xiao2021lpnet} subsequently conditions proposal generation on sentence queries and/or video context to avoid dense sampling. By contrast, another line of work~\cite{zhang20202dtan,wang2021smin,soldan2021vlgnet,gao2021ranet,zhang2021matn} takes the opposite path of enumerating all segments, yet organizes them in a 2D adjacency map to reason about their relations.

\textit{Single-stage} methods localize moments in a single shot without using proposals, and thus are often more efficient than two-stage methods. Some models decode moment boundaries from global features~\cite{yuan2019ablr,mun2020lgi,li2021cpnet,zhou2021denet,li2022visa} or learnable queries~\cite{lei2021momentdetr,woo2022lvtr,lin2023univtg}. Others densely predict onset and offset probabilities~\cite{ghosh2019excl,zhang2020vslnet,rodriguez2020tmlga,liu2021cbln,zhao2021cpn,zhang2021seqpan,nan2021ivg,zhang2022pearl}, or classify and refine pre-set anchors~\cite{chen2018tgn,zhang2019man,yuan2019scdm,wang2020cbp,liu2021ianet} at every moment. Most relevant to our work are models that represent moment candidates as points~\cite{lu2019debug,zeng2020drn,chen2020gdp,liu2022mgslnet,fang2023tcsf}. Analogous to point-based object detectors~\cite{tian2019fcos}, they classify every point in time and regress moment boundaries relative to the points. Despite their simplicity and efficiency, the performance of single-stage models lags behind two-stage models on standard benchmarks made up of short videos.

\textit{Cross-modal fusion} combines video and text information for video grounding. Early models~\cite{anne2017mcn,gao2017ctrl} fuse video and text representations right before final prediction. This simple, late-fusion approach is abandoned by subsequent models, which turn to early fusion (\eg,~\cite{xu2019qspn,zhang2019man,zhang20202dtan,zeng2020drn}) with sophisticated model design (\eg, LSTM~\cite{chen2018tgn,ghosh2019excl,xu2019qspn,wang2020cbp,liu2021cbln,zhang2022pearl}, GCN~\cite{chen2020gdp,soldan2021vlgnet,liu2022mgslnet}, memory bank~\cite{chen2019sap,liu2022mgslnet}), feature hierarchies~\cite{wang2021smin,gao2021ranet,mun2020lgi}, losses~\cite{xu2019qspn,nan2021ivg,zhang2021matn,li2022visa}, and inference procedures~\cite{zhou2021denet,zhao2021cpn}. Early fusion is believed to strengthen cross-modal reasoning, albeit at a price of model complexity. Recent Transformer-based models~\cite{zhang2021matn,lei2021momentdetr,woo2022lvtr,barrios2023localizing} inherit this early-fusion design; video features and text embeddings are concatenated at input, and attention-based feature fusion is carried out throughout the model.

\textit{Our method} demonstrates a single-stage Transformer model with a simple late fusion mechanism based on cross-attention. Late fusion allows us to amortize the cost of video processing across many sentence queries, resulting in scalable training and inference on long-form videos. Without bells and whistles, our model outperforms all prior methods on long-video benchmarks, and attains competitive performance on standard, short-video benchmarks. 

\smallskip
\noindent \textbf{Long-form video grounding.} Recent efforts~\cite{grauman2022ego4d,soldan2022mad} put forth new challenges for sentence grounding in long videos. Previous methods tailored for short videos~\cite{zhang20202dtan,zhang2020vslnet,soldan2021vlgnet} fall short on these benchmarks. They have no access to long-range video context, require dense sliding window inference, and yield unsatisfactory results. Concurrent works~\cite{hou2022cone,pan2023scanning} present multi-stage approaches with sophisticated pre-processing, ranking and training procedures for efficient long-form video grounding. However, their efficiency stems from the restrictive assumption that moments are covered by short windows. By contrast, SnAG offers a simple, single-stage solution with superior accuracy, without making assumptions about moment duration.

\smallskip
\noindent \textbf{Localizing moments in videos}. Closely related to video grounding, temporal action localization (\eg,~\cite{shou2016tal1,ma2016tal2,singh2016tal3}) aims to identify moments of actions from within a closed set of action categories. Solutions to the two tasks often evolve in tandem and share remarkably similar model design. SnAG is inspired by our prior work~\cite{zhang2022actionformer}, yet extends its design for the more challenging task of video grounding, with the goal of scaling up video grounding for long videos.

% method
\section{Scaling Up Video Grounding}

\begin{figure*}
    \centering
\includegraphics[width=1.0\linewidth]{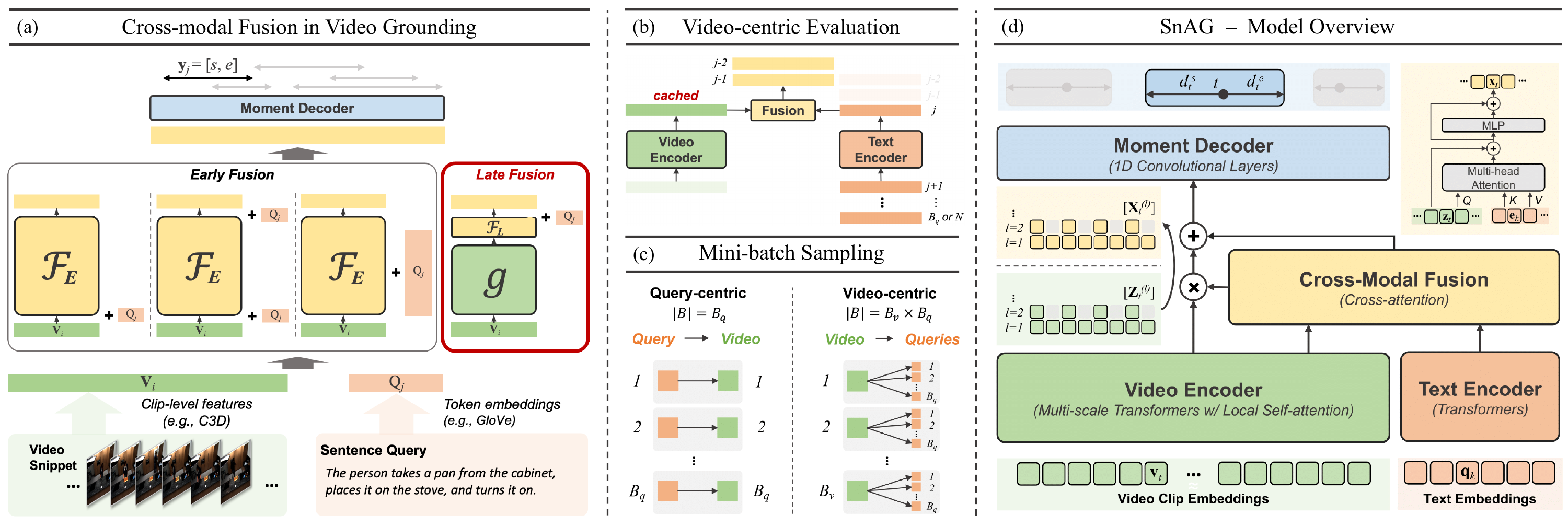}\vspace{0.5em}
    \caption{\textbf{(a)~Cross-modal fusion} is key to a video grounding model $\mathcal{F}$. Models using \emph{early fusion} jointly encode video and sentence query. SnAG revisits~\emph{late fusion} for scalable video grounding by decoupling expensive video encoding and inexpensive query encoding. \textbf{(b)~Video-centric model evaluation.} With late fusion, the output of video encoder can be cached and re-used by queries of the same video in both training and inference. \textbf{(c)~Mini-batch sampling in training}. Previous methods adopt \emph{query-centric} sampling (query~\textrightarrow~video) whereas SnAG resorts to \emph{video-centric} sampling (video~\textrightarrow~many queries) for efficient training. \textbf{(d)~Model overview}. SnAG is a simple instantiation of late fusion and video-centric training for video grounding. It separately encodes a video and its queries using Transformers, applies simple cross-attention for cross-modal fusion, and decodes moments represented as points using lightweight convolutional heads.}
    \label{fig:method}\vspace{-1em}
\end{figure*}

Our goal is to scale up video grounding for long-form videos with many queries. Denote an input video $\mathbf{V}$ as a time-indexed sequence $[\mathbf{v}_t]_{t=1}^T  =[\mathbf{v}_1,\mathbf{v}_2,...,\mathbf{v}_T]$, where $\mathbf{v}_t$ represents a short video clip centered at time $t$.\footnote{Like prior works, we consider video features and text embeddings extracted using pre-trained models (\eg, C3D~\cite{tran2015c3d} and BERT~\cite{devlin2018bert}). Yet the formulation can be easily extended to raw video frames and text tokens.} $\mathbf{V}$ is paired with $N$ sentence queries $\{\mathrm{Q}_j\}_{j=1}^N$, each describing a moment $\mathbf{y}_j=(s_j,e_j)$ in the video. 

\smallskip
\noindent \textbf{Long-form video grounding} is characterized by a great number of time steps $T$ (\eg, tens of thousands) and text queries $N$ (hundreds) for an input video $\mathbf{V}$. With a large $T$, it is no longer feasible to feed the full video $\mathbf{V}$ into a model, and thus more viable to consider a sliding window setting. Specifically, a video $\mathbf{V}$ is further partitioned into a set of $M$ snippets $\{\mathbf{V}_{i=1}^M\}$, each $\mathbf{V}_i=[\mathbf{v}_t]_{t=t_i}^{t_i + T_w}$ taken from a local time window of size $T_w$ that may overlap with others. We consider the learning of a function $\mathcal{F}$, realized in a deep network, to predict the start time $s_j$ and end time $e_j$ of the moment $\mathbf{y}_j$ given a snippet $\mathbf{V}_i$ and a query $\mathrm{Q}_j$.

\subsection{Background: Cross-Modal Fusion}\label{sec:method_background}
We review cross-modal fusion --- a key design choice of $\mathcal{F}$, and define the notation used for the rest of the paper.

\smallskip
\noindent \textbf{Early fusion.} A straightforward design is to fuse information from a video snippet $\mathbf{V}_i$ and a text query $\mathrm{Q}_j$ at input (Fig.\ \ref{fig:method}(a)). This early fusion strategy is widely adopted in prior works~\cite{zhang2019man,zhang20202dtan,zeng2020drn,zhang2021matn,lei2021momentdetr,woo2022lvtr}, and can be written as
\begin{equation}
    \mathcal{F}_{E}(\mathbf{V}_i, \mathrm{Q}_j) \mapsto \mathbf{y}_j,
\end{equation}
where $\mapsto$ defines any necessary post-processing of the model outputs (\eg, re-scaling, non-maximum suppression) in order to decode the moment $\mathbf{y}_j$.

\smallskip
\noindent \textbf{Late fusion.} Another possible design is to first embed the video snippet $\mathbf{V}_i$ and text query $\mathrm{Q}_j$, then fuse their features for grounding (Fig.\ \ref{fig:method}(a)). This late fusion strategy follows
\begin{equation}
    \mathcal{F}_{L}(g(\mathbf{V}_i), h(\mathrm{Q}_j)) \mapsto \mathbf{y}_j,
\end{equation}
where $g(\cdot)$ is the video encoder and $h(\cdot)$ the text encoder. With slight abuse of the symbols, $\mathcal{F}_{L}$ here denotes the rest part of the grounding model, including feature fusion. The key is to keep $\mathcal{F}_{L}$ lightweight, \ie, its cost should be a minor portion in comparison to the rest of the model. 

\smallskip \noindent \textbf{Design choice.} The conventional wisdom favors the design of early fusion for video grounding. Early fusion is believed to offer better interactions between visual and textual features, resulting in strong empirical results. In what follows, we analyze the computational cost of early and late fusion during inference and training, and show that proper design of late fusion allows significantly improved scalability in video duration ($M$) and number of text queries ($N$). 

\subsection{Scalable Inference}\label{sec:method_inference}

We start by analyzing the computational cost of early \vs.\ late fusion in inference. Denote the computational cost of a function by $\mathcal{C}(\cdot)$. For example, $\mathcal{C}(g)$ is the cost of the video encoder $g(\cdot)$, and $\mathcal{C}(h)$ the cost of the text encoder $h(\cdot)$.

\smallskip
\noindent \textbf{Computational cost.} We make two key observations regarding the computational cost of a video grounding model: (a) a lightweight $\mathcal{F}_{L}$ is used for late fusion, thus $\mathcal{C}(\mathcal{F}_{L}) \ll \mathcal{C}(g) + \mathcal{C}(h)$; and (b) the video encoder $g(\cdot)$ has a much higher cost than the text encoder, thus $\mathcal{C}(g) \gg \mathcal{C}(h)$. 

The first observation stems from the design of late fusion. To justify the second observation, it is important to notice that $\mathcal{C}(g)$ scales w.r.t.\ the length of input video snippet (\ie, the local window size $T_w$), whereas $\mathcal{C}(h)$ remains constant due to limited sequence length of text queries. For example, a realization of $g(\cdot)$ using a vanilla Transformer has a complexity of $\mathcal{O}(T_w^2)$. Thus, using a larger local window size naturally leads to $\mathcal{C}(g) \gg \mathcal{C}(h)$.

\smallskip 
\noindent \textbf{Cost analysis: early \vs.\ late fusion.} Evaluating an early-fusion model on a video with $M$ snippets and $N$ queries has a cost of $M N \ \mathcal{C}(\mathcal{F}_{E})$, whereas late fusion has a cost of $M N\  \mathcal{C}(\mathcal{F}_{L}) + M \mathcal{C}(g) + N \mathcal{C}(h)$. This is because an early-fusion model has to compare each snippet to every query, yet late fusion allows re-using video features for all queries of the video (\ie, \emph{video-centric} evaluation, see Fig.\ \ref{fig:method}(b)). 

Let us assume that the cost of both models can be controlled by setting $\mathcal{C}(\mathcal{F}_{E}) = \mathcal{C}(\mathcal{F}_{L}) + \mathcal{C}(g) + \mathcal{C}(h)$, \ie, evaluating either model on a pair of video snippet and sentence query has the same cost (post-processing excluded). With $\mathcal{C}(\mathcal{F}_{L}) = \alpha(\mathcal{C}(g) + \mathcal{C}(h))$ for some small $\alpha$, the ratio between the cost of early and late fusion in inference is 
\begin{align}\label{eq:cost}
\small
    \mathbf{R_{\text{inf}}} &= \frac{\mathcal{C}(\text{inference with early fusion})}{\mathcal{C}(\text{inference with late fusion})} \\
    &= \frac{M N \ \mathcal{C}(\mathcal{F}_{L}) + M N \ \mathcal{C}(g) + M N \ \mathcal{C}(h)}{M N\  \mathcal{C}(\mathcal{F}_{L}) + M\ \mathcal{C}(g) + N\ \mathcal{C}(h)} \\
    & = \frac{MN(1+\alpha)}{MN\alpha + M \mathbf{R}_{g/(g+h)} + N (1-\mathbf{R}_{g/(g+h)})}\\
    & \approx \frac{1+\alpha}{\frac{1}{N}+\alpha} \to
    \begin{cases}
		N, & \alpha\to 0 \\
        1+\frac{1}{\alpha}, & N\to \infty
    \end{cases}\label{eq:approx},
\end{align}
where $\mathbf{R}_{g/(g+h)} = \mathcal{C}(g) / (\mathcal{C}(g) + \mathcal{C}(h))$, and the approximation in Eq.\ \ref{eq:approx} is a result of $\mathbf{R}_{g/(g+h)}\approx 1$ with $\mathcal{C}(g) \gg \mathcal{C}(h)$. Eq.\ \ref{eq:approx} reveals that early fusion is $N$ times the cost of late fusion as snippet length grows ($\alpha \to 0$), and is particularly inefficient for videos with many queries ($N\to \infty$).

\subsection{Scalable Training}\label{sec:method_training}

We now show that late fusion further unlocks a novel training scheme that scales well for long-form video grounding.

\smallskip 
\noindent \textbf{Training and mini-batch sampling.}
Given a set of paired videos, text queries, and their moments $\{\mathbf{V}^{(k)}, \mathrm{Q}^{(k)}, \mathbf{y}^{(k)}\}$, training a model $\mathcal{F}$ amounts to minimizing a loss function $\mathcal{L}$ over mini-batches $B$ sampled from the dataset:
\begin{equation}
\small
    \Sigma_{(k, i, j) \in B} \ \  \mathcal{L}\left(\mathcal{F}\left(\mathbf{V}^{(k)}_i, \mathrm{Q}^{(k)}_j\right), \mathbf{y}^{(k)}_j\right).
\end{equation}

A key design largely overlooked in video grounding is the mini-batch sampling scheme during training. With multiple queries associated with a video snippet, a combination of late fusion and proper sampling scheme allows us to re-use video representations in training as is done in inference.

\smallskip 
\noindent \textbf{Query-centric sampling}. Virtually all prior works sample mini-batches uniformly at random from all pairs of valid video snippets and text queries.\footnote{A video snippet without any query is often discarded during training.} This can be interpreted as ancestral sampling: first draw a text query (and its moment), then select a video snippet covering the moment: 
\begin{equation}
\small
\begin{split}
    & \mathrm{Q}_{j'}^{(k')}, \mathbf{y}_{j'}^{(k')} \sim U \left( \{ ( \mathrm{Q}_j^{(k)}, \mathbf{y}_{j}^{(k)} ) \}_{j\times k} \right), \\
    & \mathbf{V}_{i'}^{(k')} \sim U \left( \{ (\mathbf{V}_i^{(k')}, \mathrm{Q}_{j'}^{(k')}, \mathbf{y}_{j'}^{(k')})_+ \} \right), \\
\end{split}
\end{equation}
where $U(\cdot)$ denotes a discrete uniform distribution and $\{ (\mathbf{V}_i^{(k')}, \mathrm{Q}_j^{(k')}, \mathbf{y}_{j'}^{(k')})_+ \}$ is the union of all valid triplets: a moment $\mathbf{y}_{j'}^{(k')}$ of a text query $\mathrm{Q}_{j'}^{(k')}$ in a video snippet $\mathbf{V}^{(k')}_i$.

This sampling process is repeated $|B|$ times to construct a mini-batch, with $|B|$ the batch size. With a large dataset, the resulting mini-batch will contain $|B|$ snippet-query pairs, each likely from a different video. 

\smallskip 
\noindent \textbf{Video-centric sampling}. To speed up the training of late fusion models, we propose video-centric sampling that considers every video snippet $\mathbf{V}_i^{(k)}$ as a sample. Specifically, we first draw a video based on its importance score, then sample a snippet within the video, and finally select \emph{multiple} text queries associated with the snippet. This is defined as the following ancestral sampling:
\begin{equation}
\small
\begin{split}
    & k' \sim P(k) = \frac{M^{(k)}_{+}}{\Sigma_k M^{(k)}_{+}} \quad 
    i' \sim P(i; k') = \frac{N^{(k')}_i}{\Sigma_k N^{(k')}_i} \\
    & \{ \mathrm{Q}_{j'}^{(k')}, \mathbf{y}_{j'}^{(k')} \}^{B_q} \sim U\left(\{(\mathbf{V}_{i'}^{(k')}, \mathrm{Q}_j^{(k')}, \mathbf{y}_{j}^{(k')})_+ \} \right),
\end{split}
\end{equation}
where $M^{(k)}_{+}$ denotes the number of valid snippets in a video $\mathbf{V}^{(k)}$, and $N^{(k')}_i$ the number of queries in a snippet $\mathbf{V}^{(k')}_i$. Importantly, a total of $B_q$ queries are drawn for the selected video snippet $\mathbf{V}_{i'}^{(k')}$. 

This sampling process is repeated $|B|/B_q$ times for a mini-batch, leading to $|B|/B_q$ video snippets, each likely from a different video yet with multiple ($B_q$) text queries. This mini-batch thus allows us to re-use the video representations for the sampled snippets. 

\smallskip 
\noindent \textbf{Cost analysis: query-centric \vs.\ video-centric sampling.} Fig.\ \ref{fig:method}(c) illustrates the key difference between query-centric and video-centric sampling. Similar to our analysis in Section~\ref{sec:method_inference}, it is easy to show that late fusion combined with video-centric training enables substantial speedup for long snippets ($\alpha\to 0$) and large $B_q$:
\begin{align}\label{eq:sampling}
\small
    \mathbf{R_{\text{train}}} &= \frac{\mathcal{C}(\text{late fusion + video-centric training})}{\mathcal{C}(\text{late fusion + query-centric training})} \\
    & \approx \frac{1+\alpha}{\frac{1}{B_q}+\alpha} \to
    \begin{cases}
        B_q, & \alpha\to 0 \\
		1+\frac{1}{\alpha}, & B_q\to \infty
    \end{cases}.
\end{align}

\subsection{Scalable Video Grounding: An Instantiation}\label{sec:method_model}

We now present an instantiation with late fusion and video-centric training for scalable video grounding. Our instantiation, dubbed SnAG, is shown in Fig.\ \ref{fig:method}(d). SnAG is a single-stage Transformer model where every time step represents a moment candidate. 
It combines (a) a multi-scale Transformer-based video encoder; (b) a Transformer-based text encoder; (c) cross-attentions for late fusion; and (d) convolutional heads for moment decoding. 

\smallskip 
\noindent \textbf{Video encoder $g$.} We use the ActionFormer backbone~\cite{zhang2022actionformer} as our video encoder, given its superior performance on the closely related task of temporal action localization. It transforms input video features into a multi-scale representation $[\mathbf{Z}^{(l)}]_{l=1}^L=[\mathbf{Z}^{(1)},\mathbf{Z}^{(2)},...,\mathbf{Z}^{(L)}]$ using Transformer layers with local self-attentions~\cite{choromanski2020local}. We refer readers to the original paper~\cite{zhang2022actionformer} for more details on model architecture.

\smallskip 
\noindent \textbf{Text encoder $h$.} Our text encoder is a vanilla Transformer. It takes a query $\mathrm{Q}$ with $K$ tokens $[\mathrm{q}_1,...,\mathrm{q}_K]$ and outputs their embeddings $\mathbf{E}=[\mathbf{e}_1,...\mathbf{e}_K]$. This computation is light compared to video encoding as $K\ll T_w$ for long videos. 

\smallskip 
\noindent \textbf{Cross-modal late fusion $\mathcal{F}_L$.} Our model uses a simple variant of cross-attention for fusion. Specifically, it modulates $[\mathbf{Z}^{(l)}]_{l=1}^L$ with an affine transformation given by $\mathbf{E}$:
\begin{equation}
\small
\begin{aligned}
    (\mathbf{W},\mathbf{B})^{(l)}&=\mathrm{MCA}(\mathrm{LN}(\mathbf{Z}^{(l)}),\mathrm{LN}(\mathbf{E})),
    \\
    \tilde{\mathbf{Z}}^{(l)}&=\mathbf{W}^{(l)}\odot\mathbf{Z}^{(l)}+\mathbf{B}^{(l)},
    \\
    \mathbf{X}^{(l)}&=\tilde{\mathbf{c}}\odot\mathrm{MLP}(\mathrm{LN}(\tilde{\mathbf{Z}}^{(l)}))+\tilde{\mathbf{Z}}^{(l)},
\end{aligned}
\end{equation}
where $\mathrm{LN}$ is Layer Normalization~\cite{ba2016layernorm}, $\mathrm{MLP}$ is Multi-Layer Perceptron, and $\mathrm{MCA}$ is Multi-head Cross-Attention~\cite{vaswani2017attention}, whose output is split into affine weights $\mathbf{W}^{(l)}$ and biases $\mathbf{B}^{(l)}$. $\tilde{\mathbf{c}}$ are learnable per-channel scales, and $\mathbf{X}^{(l)}$ the modulated video representations prior to moment decoding.

\smallskip 
\noindent \textbf{Moment decoding.}
After fusion, our model decodes each time step into a moment candidate, similar to previous video grounding models~\cite{lu2019debug,zeng2020drn,chen2020gdp,liu2022mgslnet}. Briefly, a classification head predicts a score $p_t^{(l)}$ for all $\mathbf{x}_t^{(l)}$ within $[\mathbf{X}^{(l)}]_{l=1}^L$, and a regression head predicts the normalized distances from $t$ to the moment boundaries $(d_t^s,d_t^e)$ if $\mathbf{x}_t^{(l)}$ is classified positive. Both heads are lightweight, each comprising three 1D convolutions. The decoded moment $\hat{\mathbf{y}}=(\hat{s},\hat{e})$ is given by
\begin{equation}
\begin{gathered}
    (t,l)=\argmax_{t,l}p_t^{(l)},
    \\
    \hat{s}=2^{l-1}\cdot(t-d_t^s),\quad\hat{e}=2^{l-1}\cdot(t+d_t^e).
\end{gathered}
\end{equation}
At inference time, we further apply Soft-NMS~\cite{bodla2017softnms} to merge overlapping moment predictions.

\smallskip \noindent \textbf{Training and loss function.} We use video-centric training as described in Section~\ref{sec:method_training}. The loss function for a pair of video snippet and text query is the sum of two terms: a Focal loss~\cite{lin2017focal} $\mathcal{L}_{cls}$ for moment classification and a Distance-IoU loss~\cite{zheng2020diou} $\mathcal{L}_{reg}$ defined on positive time steps for distance regression. Formally,
\begin{equation}\label{eq:loss}
\small
    \mathcal{L}=\frac{1}{C}(\mathcal{L}_{cls}+\lambda_{reg}\mathds{1}_t^{(l)}\mathcal{L}_{reg}),
\end{equation}
where $\mathds{1}_t^{(l)}$ is an indicator function that evaluates to one on positive time steps, $C=\sum_{t, l}\mathds{1}_t^{(l)}$ is the total number of positive samples within the snippet, and $\lambda_{reg}$ balances the two loss terms. Center sampling~\cite{zhang2022actionformer} is used for selecting positive time steps around the center of a target moment. 

% experiments
\section{Experiments}

\begin{figure}
\centering
\includegraphics[width=0.72\linewidth]{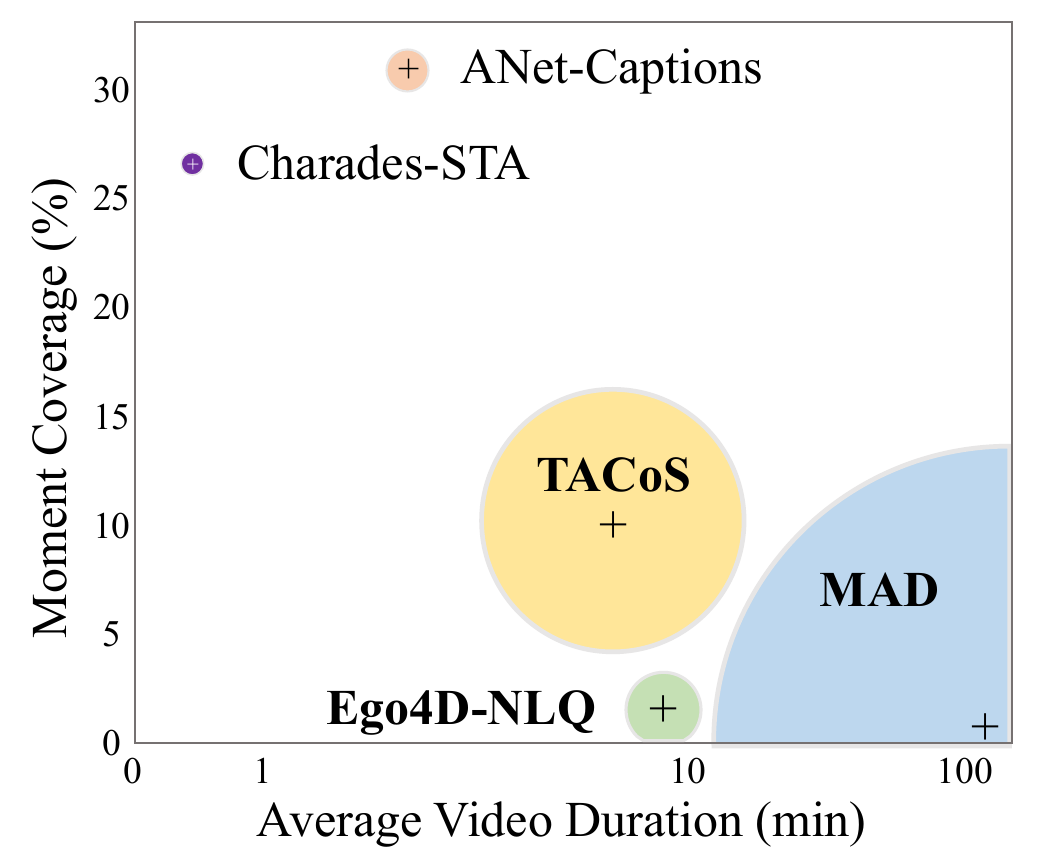}
\vspace{0.5em}
\caption{\textbf{Visualization of dataset statistics.} Circle radius is in proportion to average number of queries per video. Long-video benchmarks (Ego4D-NLQ, MAD and TACoS) consist of longer videos with more queries and exhibit low moment coverage compared to short-video benchmarks (ANet-Captions and Charades).}\label{fig:data_stats}
\end{figure}

% Ego4D results
\begin{table}[th]
\begin{center}
\resizebox{\linewidth}{21mm}
{
    \begin{tabular}{llcccccc} 
    \toprule[1.5pt]
    % header
    \multirow{2}{*}{}       
    & \multirow{2}{*}{Model} 
    & \multicolumn{3}{c}{R@1}                                       
    & \multicolumn{3}{c}{R@5} \\ 
    \cmidrule(r){3-5} \cmidrule(r){6-8}
    & & 0.3 & 0.5 & Avg & 0.3 & 0.5 & Avg \\ 
    \midrule[0.5pt]
    % block 1
    % \rowcolor was supposed to be used at the begining of row, and will throw the nonalign error here. 
    % Switching to \cellcolor for individual cells solves this error. 
    \multirow{6}{*}{\begin{turn}{-270}\makecell{SF + BERT}\end{turn}} 
    & 2D-TAN~\cite{zhang20202dtan} & 5.04 & 2.02 & 3.53 & 12.89 & 5.88 & 9.39 \\
    & VSLNet~\cite{zhang2020vslnet} & 5.45 & 3.12 & 4.29 & 10.74 & 6.63 & 8.69 \\
    & CONE~\cite{hou2022cone} & \textbf{10.40} & 5.03 & 7.72 & 22.74 & 11.87 & 17.31 \\
    & SOONet~\cite{pan2023scanning} & 8.00 & 3.76  & 5.88 & 22.40 & 11.09 & 16.75 \\  
    \specialrule{0em}{1pt}{1pt}
    & 
    \cellcolor{black!5} \textbf{SnAG}~\footnotesize{(video-centric)} &  \cellcolor{black!5} \underline{9.83} &  \cellcolor{black!5} \underline{6.83} &  \cellcolor{black!5} \underline{8.33} &  \cellcolor{black!5} \textbf{27.93} &  \cellcolor{black!5} \textbf{19.27} &  \cellcolor{black!5} \textbf{23.60} \\
    &\cellcolor{black!5} \textbf{SnAG}~\footnotesize{(query-centric)} & \cellcolor{black!5} \textbf{10.43} & \cellcolor{black!5} \textbf{7.15} & \cellcolor{black!5} \textbf{8.79} & \cellcolor{black!5} \underline{27.59} & \cellcolor{black!5} \textbf{19.33} & \cellcolor{black!5} \underline{23.46} \\ 
    \midrule[0.5pt]
    % block 2
    \multirow{4}{*}{\begin{turn}{-270}EgoVLP\end{turn}}     
    & VSLNet~\cite{zhang2020vslnet} & 10.84 & 6.81 & 8.83 & 18.84 & 13.45 & 16.15 \\
    & CONE~\cite{hou2022cone} & 14.15 & 8.18 & 11.17 & 30.33 & 18.02 & 24.18 \\
    \specialrule{0em}{1pt}{1pt}
    &\cellcolor{black!5} \textbf{SnAG}~\footnotesize{(video-centric)} & \cellcolor{black!5} \underline{15.72} & \cellcolor{black!5} \underline{10.78} & \cellcolor{black!5} \underline{13.25} & \cellcolor{black!5} \textbf{38.39} & \cellcolor{black!5} \textbf{27.44} & \cellcolor{black!5} \textbf{32.92} \\
    &\cellcolor{black!5} \textbf{SnAG}~\footnotesize{(query-centric)} & \cellcolor{black!5} \textbf{15.87} & \cellcolor{black!5} \textbf{11.26} & \cellcolor{black!5} \textbf{13.57} & \cellcolor{black!5} \underline{38.26} & \cellcolor{black!5} \underline{27.16} & \cellcolor{black!5} \underline{32.71} \\ 
    \bottomrule[1.5pt]
    \end{tabular}
}
\vspace{0.5em}
\caption{\textbf{Results on Ego4D-NLQ.} SnAG outperforms all baselines. Best results are in \textbf{bold} and second best \underline{underlined}.}\label{table:results_ego4d}\vspace{-1em}
%Best results are in \textbf{bold} and second best \underline{underlined}. }
%SnAG outperforms all previous baselines and the concurrent method of CONE by a significant margin.
\end{center}
\end{table}

Our main experiments include extensive comparisons of SnAG to strong baselines on five challenging benchmarks for video grounding. We further provide ablations on cross-modal fusion and analyze the efficiency of SnAG in both training and inference. More results are in the supplement.

\smallskip \noindent \textbf{Datasets.} We categorize five benchmark datasets into two groups based on their video and query statistics (Fig.~\ref{fig:data_stats}).

\smallskip \textit{Long videos, many queries.} Ego4D-NLQ~\cite{grauman2022ego4d}~\footnote{All experiments are conducted using the Ego4D \emph{v1} dataset.} is a large collection of egocentric videos on daily human activities. The videos are 3.5 to 20 minutes long with an average of 11.6 queries. MAD~\cite{soldan2022mad} contains 1.2K hours of movies with 384K queries transcribed from audio description. The videos are 47 to 202 minutes long. TACoS~\cite{regneri2013tacos} is a conventional benchmark with 10.1 hours of cooking videos, with an average of 143.5 queries per video. These datasets expose the scalability challenges that motivate our work and thus are the main focus of our experiments.

% MAD results
\begin{table*}[]
\resizebox{0.9\linewidth}{20mm}{
\begin{tabular}{lcccccccccccc}
\toprule[2pt]
\multirow{2}{*}{Model} & \multicolumn{3}{c}{R@1} & \multicolumn{3}{c}{R@5} & \multicolumn{3}{c}{R@10} & \multicolumn{3}{c}{R@50} \\ \cmidrule(r){2-4} \cmidrule(r){5-7} \cmidrule(r){5-7} \cmidrule(r){8-10} \cmidrule(r){11-13}
& 0.1  & 0.3  & 0.5  & 0.1   & 0.3   & 0.5  & 0.1   & 0.3   & 0.5   & 0.1   & 0.3   & 0.5 \\
\midrule[0.5pt]
CLIP~\cite{radford2021clip}    & 6.57 & 3.13 & 1.39 & 15.05 & 9.85  & 5.44 & 20.26 & 14.13 & 8.38  & 37.92 & 28.71 & 18.80 \\
VLG-Net~\cite{soldan2021vlgnet} & 3.64 & 2.76 & 1.65 & 11.66 & 9.31  & 5.99 & 17.39 & 14.56 & 9.77  & 39.78 & 34.27 & 24.93 \\
Moment-DETR~\cite{lei2021momentdetr} & 0.31 & 0.24 & 0.16 & 1.52 & 1.14 & 0.28 & 2.79 & 2.06 & 1.20 & 11.08 & 7.97 & 4.71 \\ 
CONE~\cite{hou2022cone}    & 8.90 & 6.87 & 4.10 & 20.51 & 16.11 & 9.59 & 27.20 & 21.53 & 12.82 & 43.36 & 34.73 & 20.56 \\
SOONet~\cite{pan2023scanning} & \textbf{11.26} &\textbf{9.00 }& 5.32 & 23.21 & 19.64 &13.14 & 30.36 & 26.00 & 17.84 &50.32 & 44.78 & 32.59 \\
\specialrule{0em}{1pt}{1pt}
\rowcolor{black!5} \textbf{SnAG}~\footnotesize{(video-centric)}     & \uline{10.28}    & \uline{8.46}    & \textbf{5.55}    & \textbf{24.42}     & \textbf{20.60}     & \textbf{13.75}    & \textbf{32.23}     & \textbf{27.50}     & \textbf{19.00}     & \textbf{52.28}     & \textbf{46.68}     & \textbf{35.24}     \\
\rowcolor{black!5} \textbf{SnAG}~\footnotesize{(query-centric)}  & 10.18   & 8.37    & \underline{5.44}    & \underline{24.30}     & \underline{20.35}     & \underline{13.51}    & \underline{32.05}    & \underline{27.25}     & \underline{18.75}     &   \underline{52.05}   & \underline{46.53}  & \underline{34.83}\\ 
\bottomrule[2pt]
\end{tabular}
}
\centering
\vspace{0.5em}
\caption{\textbf{Results on MAD.} SnAG outperforms all baselines. Best results are in \textbf{bold} and second best \underline{underlined}.}
\label{table:results_mad}\vspace{-1em}
\end{table*}
% & 10.62   & 8.71    & 5.73    & 24.55     & 20.70     & 14.06    & 32.16    & 27.64     & 19.36     &   51.38   & 46.11  & 34.91

\begin{figure}
\centering
\includegraphics[width=1.0\linewidth]{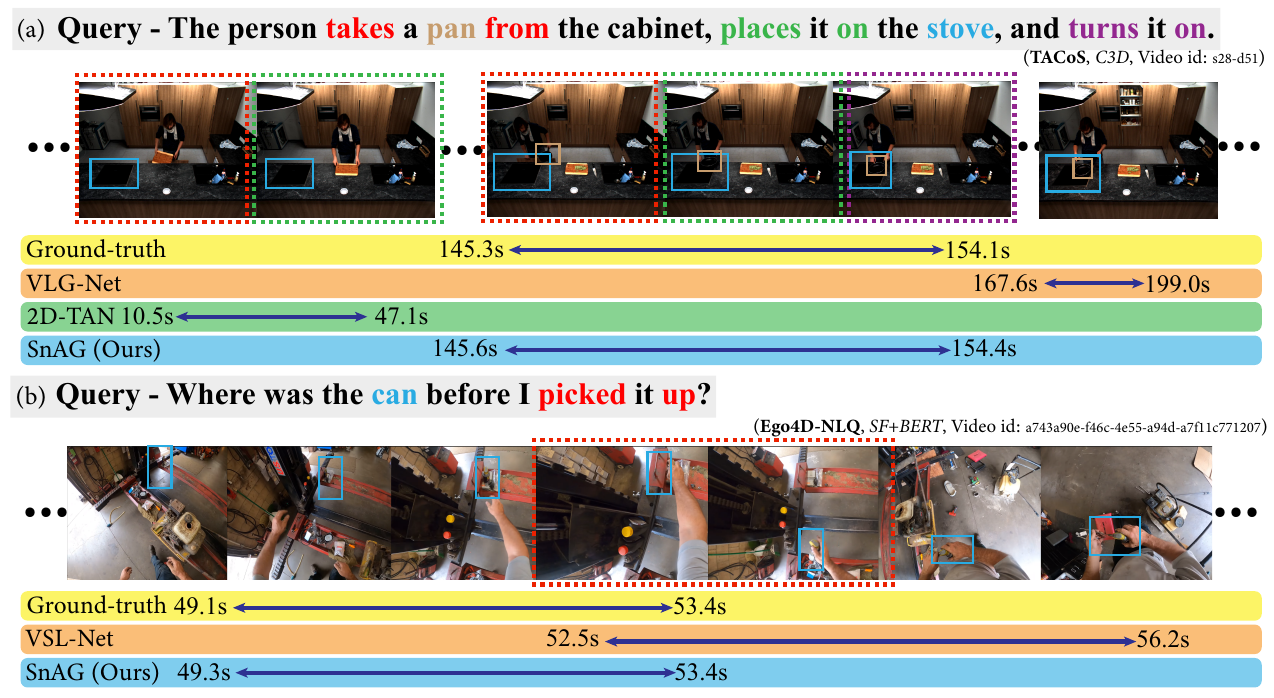}
\vspace{-0.5em}
\caption{\textbf{Visualization of moment predictions.} SnAG can \textbf{(a)} comprehend complex text queries with multiple objects and actions; \textbf{(b)} reason about temporal ordering of events.
}\vspace{-1.5em}
\label{fig:viz}
\end{figure}

\smallskip \textit{Short videos, few queries.} 
Charades-STA~\cite{sigurdsson2016charades} is an action recognition dataset later annotated for video grounding~\cite{gao2017ctrl}. The videos are $\sim$30 seconds long with $\sim$2.4 queries per video. ActivityNet-Captions~\cite{krishna2017activitynet} was collected for dense video captioning and later repurposed for video grounding. The videos are two minutes long on average with a small set of $\sim$3.65 queries. Several works~\cite{otani2020uncover,yuan2021closer} advise against using these datasets as benchmarks for the temporal biases in their moment annotations. Nevertheless, we report results on these datasets for completeness.

\smallskip \noindent \textbf{Baselines.} On TACoS, Charades and ANet-Captions, we compare to strong baselines including two-stage~\cite{xu2019qspn,xiao2021bpnet,liu2021apgn} and single-stage~\cite{wang2021smin,zhou2021denet,liu2021ianet} models with various fusion operations (LSTM~\cite{wang2020cbp,liu2021cbln}, GCN~\cite{soldan2021vlgnet,liu2022mgslnet}, co-attention~\cite{li2021cpnet,zhao2021cpn}, \etc). On Ego4D and MAD, we compare to their official baselines (2D-TAN~\cite{zhang20202dtan} and VSLNet~\cite{zhang2020vslnet} for Ego4D, CLIP~\cite{radford2021clip} and VLG-Net~\cite{soldan2021vlgnet} for MAD) as well as the concurrent method of CONE~\cite{hou2022cone} and SOONet~\cite{pan2023scanning} that similarly address scalable video grounding. We discuss additional baselines in the supplement.

\smallskip \noindent \textbf{Evaluation metric.} We report Recall$@k$ at various temporal intersection-over-union threshold $\theta$ (R@$k$,~tIoU=$\theta$) for all datasets. It measures the percentage of sentence queries with at least one of the top-$k$ moment predictions whose temporal overlap with the ground-truth moment exceeds $\theta$.

\subsection{Long Videos, Many Queries}

\noindent \textbf{Experiment setup.} 
For Ego4D, We consider (1) SlowFast~\cite{feichtenhofer2019slowfast} video features with BERT~\cite{devlin2018bert} text features, and (2) EgoVLP~\cite{lin2022egovlp} video and text features. We train on the \textit{train} split and report R@\{1,~5\}, tIoU=\{0.3,~0.5\} on the \textit{val} split. For MAD, we use CLIP~\cite{radford2021clip} features for both videos and text. We train on the \textit{train} split and report R@\{1,~5,~10,~50\}, tIoU=\{0.1,~0.3,~0.5\} on the \textit{test} split. For TACoS, we use C3D video features~\cite{tran2015c3d} and GloVe~\cite{pennington2014glove} for sentence embeddings. We train on the \textit{train} split and report R@\{1,~5\}, tIoU=\{0.5,~0.7\} on the \textit{test} split. We report results using both video-centric and query-centric training on all datasets.

\smallskip \noindent \textbf{Results on Ego4D-NLQ} (Table~\ref{table:results_ego4d}). SnAG sets the new state of the art, achieving mean R@1 / R@5 scores of 8.79\% / 23.60\% with SlowFast/BERT features, and 13.57\% / 32.92\% with EgoVLP features, doubling or even tripling the results of the official baselines~\cite{zhang20202dtan,zhang2020vslnet}. Compared to the strong baselines of CONE~\cite{hou2022cone} and SOONet~\cite{pan2023scanning}, SnAG improves mean R@1 and R@5 respectively by more than 1.05 and 6.29 absolute percentage points without taking advantage of external features.\footnote{CONE~\cite{hou2022cone} leverages external CLIP~\cite{radford2021clip} features in several steps.} Further, SnAG demonstrates a larger performance gain on the more stringent tIoU threshold of 0.5, indicating more precise moment localization.

\smallskip \noindent \textbf{Results on MAD} (Table~\ref{table:results_mad}). 
SnAG sets new records on the challenging MAD dataset. It improves over CLIP~\cite{radford2021clip} and VLG-Net~\cite{soldan2021vlgnet} on mean R@1 / R@5 at tIoU=0.5 by 3.90 / 7.76 absolute percentage points. In the meantime, it outperforms CONE~\cite{hou2022cone} and SOONet~\cite{pan2023scanning}, with larger gain for more stringent tIoU thresholds. Further, SnAG achieves stronger results as input snippet length increases, reaching the best accuracy with $T_w=$16384 (Supplement Fig.~\ref{fig:snippet_len}). This scaling behavior highlights the strength of our scalable model design for long-form video grounding.

% ablation results
\begin{table}[th]
\resizebox{0.9\linewidth}{35mm}{
\begin{tabular}{lcccc}
\toprule[1.5pt]
\multirow{2}{*}{Model} & \multicolumn{2}{c}{R@1} & \multicolumn{2}{c}{R@5} \\ \cmidrule{2-3} \cmidrule{4-5}
                           & 0.3        & 0.5        & 0.3        & 0.5        \\
\midrule
\multicolumn{5}{c}{(a)~Fusion stage (early \vs.\ late fusion)}                                               \\
\midrule
No fusion                  & 10.85      & 7.85       & 46.46      & 31.19      \\
\specialrule{0em}{2pt}{2pt}
SnAG-E                     & 53.34      & 44.01      & 79.56      & 69.41      \\
SnAG-EL                    & 54.84      & 44.54      & \textbf{81.13}      & 70.11      \\
SnAG-M                     & 51.41      & 40.79      & 77.56      & 65.71      \\
\specialrule{0em}{2pt}{2pt}
\rowcolor{black!5}\textbf{SnAG~(ours)} & \textbf{55.14}      & \textbf{44.94}      & \textbf{81.10}       & \textbf{70.66}     \\
\midrule
\multicolumn{5}{c}{(b)~Fusion operation (global \vs.\ token-wise)}                                           \\
\midrule
Cat                        & 49.39      & 39.24      & 77.31      & 65.36      \\
Add                        & 51.91      & 41.81      & 77.98      & 66.51      \\
Scale                      & 55.01      & 43.96      & 80.63      & 70.11      \\
Add + Scale                     & 53.36      & 43.09      & 80.43      & 69.58      \\
\specialrule{0em}{2pt}{2pt}
XAttn                      & 54.84      & \textbf{45.06}      & 79.96      & 70.08      \\
\rowcolor{black!5}\textbf{XAttn-affine~(ours)} & \textbf{55.14}      & 44.94      & \textbf{81.10}      & \textbf{70.66}      \\
% \midrule
% \multicolumn{5}{c}{(c)~Number of Fusion Steps}                                            \\
% \midrule
% 1                          & 55.14      & 44.94      & 81.10      & \textbf{70.66}      \\
% \rowcolor{black!5}\textbf{2~(ours)}      & \textbf{56.44}      & 44.86      & 81.15      & \textbf{70.66}      \\
% 3                          & 55.81      & \textbf{45.61}      & 81.03      & 70.28      \\
% 4                          & 54.44      & 43.91      & \textbf{81.65}      & 70.46      \\
\bottomrule[1.5pt]
\end{tabular}
}
\centering
\vspace{0.5em}
\caption{\textbf{Ablations results on TACoS.} \textbf{(a)} Late fusion yields comparable results to early fusion (E, EL, M); \textbf{(b)} Fusion with token-wise sentence embeddings (XAttn) produces better results than fusion with global sentence embeddings (Cat, Add, Scale).
%; \emph{(c)} our method using various number of fusion steps.
}
\label{table:ablations}\vspace{-1em}
\end{table}

\smallskip \noindent \textbf{Results on TACoS} (Table~\ref{table:results_tacos_charades_anet} (left)). SnAG achieves an R@1 / R@5 of 44.86\% / 70.66\% at tIoU=0.5, outperforming the strongest baseline of MATN~\cite{zhang2021matn} by a significant margin (+7.5\% R@1, +12.1\% R@5). This confirms that our simple single-stage design with late fusion is a strong alternative to multi-stage models with sophisticated fusion mechanisms. Importantly, our scalable design enables learning from full-resolution video features, whereas the baselines often aggressively reduce feature resolution to contain their training and inference cost on long-form videos. We hypothesize that full-resolution video features capture detailed event dynamics and are key to the strong performance of SnAG.

\smallskip \noindent \textbf{Result visualization.} 
We visualize predictions of SnAG and the baselines on TACoS and Ego4D-NLQ videos.\footnote{MAD videos are not publicly available due to copyright issues.} In Fig.\ \ref{fig:viz}(a), a long query describes the complex interplay between two objects (pan and stove) and three actions (``take from", ``place on", and ``turn on"). VLG-Net~\cite{soldan2021vlgnet} recognized both objects yet none of the actions, while 2D-TAN~\cite{zhang20202dtan} identified one action (``take from") but at the same time missed one object (pan). By contrast, SnAG followed the query semantics and produced accurate localization. The other query in Fig.\ \ref{fig:viz}(b) tests the understanding of temporal context. SnAG leveraged the keyword ``\emph{before}" to identify the correct moment before the action (``pick up"), whereas VSLNet~\cite{zhang2020vslnet} only attended to the object (can), thus mistakenly predicting a moment \emph{after} the action.

\subsection{Short Videos, Few Queries}

% TACoS, Charades and ANet results
\begin{table*}[t]
\vspace{-0.5cm}
\resizebox{\linewidth}{42mm}{
\begin{tabular}{l||cccc|cccc|cccc|cccc}
\toprule[1.5pt]
\multirow{3}{*}{Method} &
  \multicolumn{4}{c|}{\emph{TACoS~(C3D)}} &
  \multicolumn{4}{c|}{\emph{Charades-STA~(I3D)}} &
  \multicolumn{4}{c|}{\emph{Charades-STA~(C3D)}} &
  \multicolumn{4}{c}{\emph{ActivityNet-Captions~(C3D)}} \\ \cline{2-17}
 &
  \multicolumn{2}{c}{R@1} &
  \multicolumn{2}{c|}{R@5} &
  \multicolumn{2}{c}{R@1} &
  \multicolumn{2}{c|}{R@5} &
  \multicolumn{2}{c}{R@1} &
  \multicolumn{2}{c|}{R@5} &
  \multicolumn{2}{c}{R@1} &
  \multicolumn{2}{c}{R@5} \\ \cline{2-17} 
 & 0.3   & 0.5  & 0.3   & 0.5   & 0.5   & 0.7   & 0.5   & 0.7   & 0.5   & 0.7   & 0.5   & 0.7   & 0.5   & 0.7   & 0.5   & 0.7   \\
 \midrule[0.5pt]
CTRL~\footnotesize{\textcolor{black!50}{ICCV 17~\cite{gao2017ctrl}} }     & 18.32 & 13.30 & 36.69 & 25.42 & -     & -     & -     & -     & 23.63 & 8.89  & 58.92 & 29.52 & -     & -     & -     & -     \\
QSPN~\footnotesize{\textcolor{black!50}{AAAI 19~\cite{xu2019qspn}} }     & -     & -     & -     & -     & -     & -     & -     & -     & 35.60 & 15.80 & 79.40 & 45.40 & 27.70 & 13.60 & 59.20 & 38.30 \\
SCDM~\footnotesize{\textcolor{black!50}{NeurIPS 19~\cite{yuan2019scdm}}} & 27.64 & 23.27 & 40.06 & 33.49 & 54.92 & 34.26 & 76.50 & 60.02 & -     & -     & -     & -     & 36.90 & 20.28 & 66.84 & 42.92 \\
2D-TAN~\footnotesize{\textcolor{black!50}{AAAI 20~\cite{zhang20202dtan}}}  & 45.61 & 35.77 & 69.11 & 57.31 & 56.64 & 36.21 & 89.14 & 61.13 & 41.10 & 23.25 & 81.53 & 48.55 & 46.16 & 29.21 & 78.80 & 60.85 \\
CBP~\footnotesize{\textcolor{black!50}{AAAI 20~\cite{wang2020cbp}}}      & 27.31 & 24.79 & 43.64 & 37.40 & -     & -     & -     & -     & 36.80 & 18.87 & 70.94 & 50.19 & 35.76 & 17.80 & 65.89 & 46.20 \\
DRN~\footnotesize{\textcolor{black!50}{CVPR 20~\cite{zeng2020drn}}}      & -     & 23.17 & -     & 33.36 & 53.09 & 31.75 & 89.06 & 60.05 & 45.40 & 26.40 & 88.01 & 55.38 & 45.45 & 24.36 & 77.97 & 50.30 \\
BPNet~\footnotesize{\textcolor{black!50}{AAAI 21~\cite{xiao2021bpnet}}}    & 25.96 & 20.96 & -     & -     & 50.75 & 31.64 & -     & -     & 38.25 & 20.51 & -     & -     & 42.07 & 24.69 & -     & -     \\
CPNet~\footnotesize{\textcolor{black!50}{AAAI 21~\cite{li2021cpnet}}}    & 42.61 & 28.29 & -     & -     & 60.27 & 38.74 & -     & -     & 40.32 & 22.47 & -     & -     & 40.56 & 21.63 & -     & -     \\
SMIN~\footnotesize{\textcolor{black!50}{CVPR 21~\cite{wang2021smin}}}     & 48.01 & 35.24 & 65.18 & 53.36 & 64.06 & 40.75 & 89.49 & 68.09 & -     & -     & -     & -     & 48.46 & 30.34 & 81.16 & 62.11 \\
CBLN~\footnotesize{\textcolor{black!50}{CVPR 21~\cite{liu2021cbln}}}     & 38.98 & 27.65 & 73.12 & 46.24 & 61.13 & 38.22 & 90.33 & 61.69 & 47.94 & 28.22 & 88.20 & 57.47 & 48.12 & 27.60 & 79.32 & 63.41 \\
CPN~\footnotesize{\textcolor{black!50}{CVPR 21~\cite{zhao2021cpn}}}      & 47.69 & 36.33 & -     & -     & 51.07 & 31.54 & -     & -     & -     & -     & -     & -     & 45.10 & 28.10 & -     & -     \\
DeNet~\footnotesize{\textcolor{black!50}{CVPR 21~\cite{zhou2021denet}}}    & -     & -     & -     & -     & 59.70 & 38.52 & 91.24 & 66.83 & -     & -     & -     & -     & 43.79 & -     & 74.13 & -     \\
MATN~\footnotesize{\textcolor{black!50}{CVPR 21~\cite{zhang2021matn}}}     & 48.79 & 37.57 & 67.63 & 57.91 & -     & -     & -     & -     & -     & -     & -     & -     & 48.02 & 31.78 & 78.02 & 63.18 \\
VLG-Net~\footnotesize{\textcolor{black!50}{ICCVW 21~\cite{soldan2021vlgnet}}} & 45.46 & 34.19 & 70.38 & 56.56 & -     & -     & -     & -     & -     & -     & -     & -     & 46.32 & 29.82 & 77.15 & 63.33 \\
APGN~\footnotesize{\textcolor{black!50}{EMNLP 21~\cite{liu2021apgn}}}    & 40.47 & 27.86 & 59.98 & 47.12 & 62.58 & 38.86 & 91.24 & 62.11 & 48.20 & 29.37 & 89.05 & 58.49 & 48.92 & 28.64 & 78.87 & 63.19 \\
IA-Net~\footnotesize{\textcolor{black!50}{EMNLP 21~\cite{liu2021ianet}}}  & 37.91 & 26.27 & 57.62 & 46.39 & 61.29 & 37.91 & 89.78 & 62.04 & -     & -     & -     & -     & 48.57 & 27.95 & 78.99 & 63.12 \\
RaNet~\footnotesize{\textcolor{black!50}{EMNLP 21~\cite{gao2021ranet}}}    & 43.34 & 33.54 & 67.33 & 55.09 & 60.40 & 39.65 & 89.57 & 64.54 & - & - & - & - & 45.59 & 28.67 & 75.93 & 62.97 \\
MGSL-Net~\footnotesize{\textcolor{black!50}{AAAI 22~\cite{liu2021ianet}}} & 42.54 & 32.27 & 63.39 & 50.13 & 63.98 & 41.03 & \uline{93.21} & 63.85 & -     & -     & -     & -     & \uline{51.87} & 31.42 & \uline{82.60} & 66.71 \\
MMN~\footnotesize{\textcolor{black!50}{AAAI 22~\cite{wang2022mmn}}} & 39.24 & 26.17 & 62.03 & 47.39 & - & - & - & - & -     & -     & -     & -     & 48.59 & 29.26 & 79.50 & 64.76 \\
SSRN~\footnotesize{\textcolor{black!50}{EMNLP 22~\cite{zhu2023ssrn}}} & 45.10 & 34.33 & 65.26 & 51.85 & \textbf{65.59} & 42.65 & \textbf{94.76} & 65.48 & 50.39     & 31.42     & \uline{90.68}     & 59.94     & \textbf{54.49} & \uline{33.15} & \textbf{84.72} & \textbf{68.48} \\
G2L~\footnotesize{\textcolor{black!50}{ICCV 23~\cite{li2023g2l}}} & 42.74 & 30.95 & 65.83 & 49.86 & - & - & - & - & -     & -     & -     & -     & 51.68 & \textbf{33.35} & 81.32 & \uline{67.60} \\
% TCSF~\footnotesize{\textcolor{black!50}{CVPR 23~\cite{fang2023tcsf}}} & \textcolor{black!50}{49.82} & \textcolor{black!50}{38.53} & \textcolor{black!50}{68.60} & \textcolor{black!50}{59.89} & \textcolor{black!50}{53.85} & \textcolor{black!50}{37.20} & \textcolor{black!50}{90.86} & \textcolor{black!50}{58.95} & - & - & - & - & \textcolor{black!50}{66.87} & \textcolor{black!50}{48.38} & \textcolor{black!50}{88.75} & \textcolor{black!50}{80.24} \\
\rowcolor{black!5} \textbf{SnAG}~\footnotesize{(video-centric)}          & \textbf{56.44} & \textbf{44.86} & \textbf{81.15} & \textbf{70.66} & 64.62 & \textbf{46.26} & 92.55 & \textbf{71.94} & \textbf{51.72} & \textbf{33.52} & \textbf{92.55} & \textbf{64.11} & 48.55 & 30.56 & 81.71 & 63.41 \\
\rowcolor{black!5} \textbf{SnAG}~\footnotesize{(query-centric)}          & \underline{55.01} & \uline{44.51} & \uline{80.68} & \uline{70.13} & \uline{65.13} & \uline{45.91} & 92.80 & \uline{71.75} & \uline{51.10} & \uline{31.77} & 90.22 & \uline{62.15} & 47.55 & 29.65 & 81.55 & 62.21\\
\bottomrule[1.5pt]
\end{tabular}
}
\centering
\vspace{0.5em}
\caption{\textbf{Results on TACoS, Charades-STA and ActivityNet-Captions.} SnAG outperforms all baselines on TACoS and Charade-STA by a large margin, while being highly competitive on ActivityNet-Captions. Best results are in \textbf{bold} and second best \underline{underlined}.}
\label{table:results_tacos_charades_anet}
% \vspace{-0.5em}
\end{table*}

\smallskip \noindent \textbf{Experiment setup.} We extract video features using both C3D~\cite{tran2015c3d} (pretrained on Sports-1M~\cite{karpathy2014sports1m}) and I3D~\cite{carreira2017i3d} (pretrained on Kinetics~\cite{kay2017kinetics}) for Charades-STA, and use the official C3D features for ActivityNet-Captions. We use GloVe word embeddings~\cite{pennington2014glove} to represent sentence queries for both datasets. We train on their respective \textit{train} splits, report R@\{1,~5\}, tIoU=\{0.5,~0.7\} on the \textit{test} split of Charades-STA, and R@\{1,~5\}, tIoU=\{0.3,~0.5\} on the \textit{val\_2} split of ActivityNet-Captions. Following standard practice, we resize all video features to a uniform length of 256.

\smallskip \noindent \textbf{Results on Charades-STA} (Table~\ref{table:results_tacos_charades_anet} (middle)). SnAG outperforms all previous methods by a wide margin. It obtains 33.52\% R@1 and 64.11\% R@5 at tIoU=0.7 with C3D features, 2.10 and 4.52 absolute percentage points higher than the latest SSRN~\cite{zhu2023ssrn} baseline. It further attains 46.26\% R@1 and 71.94\% R@5 at tIoU=0.7 with I3D features, setting new state of the art on this competitive benchmark.

\smallskip \noindent \textbf{Results on ActivityNet-Captions} (Table~\ref{table:results_tacos_charades_anet} (right)). SnAG achieves competitive results, reaching an R@1 / R@5 score of 30.56\% / 63.41\%  at tIoU=0.7. It is only outperformed by a few latest models~\cite{liu2022mgslnet,zhu2023ssrn,li2023g2l} while being much simpler.

\subsection{Ablation Studies}

We conduct extensive ablations on TACoS to study key design choices for cross-modal fusion.

\smallskip \noindent \textbf{Early \vs.\ late fusion.} Does late fusion enable better scalability at the expense of accuracy? We study three variants of SnAG with common early-fusion designs: (1)~\emph{SnAG-E} with fusion before the pyramid; (2)~\emph{SnAG-EL} with fusion before and after the pyramid; (3)~\emph{SnAG-M} with repeated fusion within the pyramid. Table~\ref{table:ablations} confirms that early and late fusion yield comparable results.

\begin{figure*}[!t]
\centering
\includegraphics[width=1.0\linewidth]{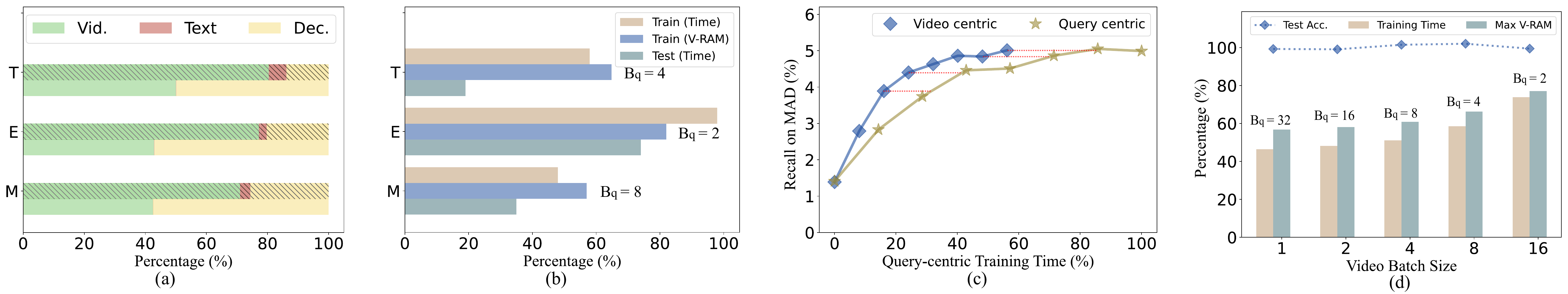}
\vspace{-1em}
\caption{\textbf{(a) Model capacity.} T, E, M denotes TACoS, Ego4D and MAD, respectively (also in (b)). Shaded bar: \% of parameters, Colored bar: \% of MACs. SnAG places more parameters and compute in video encoder for less expensive per-query evaluation (fusion~+~decoding).  \textbf{(b, c) Video-centric \vs query-centric inference (b) and training (c).} SnAG saves up to 50\% training time, 40\% GPU memory and 80\% test time, and delivers up to 40\% faster convergence relative to query-centric training~/~inference. \textbf{(d) Effect of $B_q$ on training efficiency and test accuracy on MAD.} Training on MAD is faster and takes less GPU memory as $B_q$ grows, while the test accuracy is unaffected.}
\label{fig:efficiency}
\vspace{-0.5em}
\end{figure*}

\smallskip \noindent \textbf{Choice of fusion operation.} We study fusion operations used by strong baselines, including point-wise addition, multiplication, and concatenation of video representations with global sentence embeddings, as well as vanilla cross-attention with token-level sentence embeddings. According to Table~\ref{table:ablations}, token-level embeddings produce better results, and our cross-attention variant further boosts performance.

\subsection{Efficiency Analysis}

Finally, we conduct an efficiency analysis of our method.

\smallskip \noindent \textbf{Breakdown of model capacity.} A key design choice of SnAG is to carefully balance the model capacity among its video encoder, text encoder, and moment decoder including the fusion module and the localization heads. For a detailed breakdown of model capacity, we measure their respective parameter size and MACs in percentage. Fig.\ \ref{fig:efficiency}(a) indicates that SnAG's video encoder absorbs a significant portion of parameters and compute, such that per-query evaluation (\ie, text encoding, fusion, and decoding) occupies $\sim$50\% of the total computation. This design is naturally in favor of video-centric training and inference, where the costly video encoding is shared among many queries. 

\smallskip \noindent \textbf{Video-centric \vs.\ query-centric training.} Fig.\ \ref{fig:efficiency}(b) presents the training time and GPU memory usage of video-centric training relative to conventional query-centric training. The results show that the amount of saving correlates with snippet length and number of queries. Video-centric training is thus preferred for long videos with many queries.

\smallskip \noindent \textbf{$B_q$ in video-centric training.} We further study how $B_q$, the number of text query samples per video, trades off accuracy for efficiency in video-centric training. On MAD, we fix the batch size $|B|=32$ for a mini-batch $B$, vary $B_q$ between 1 and $|B|$, and report training time, GPU memory usage and test accuracy relative to query-centric training (\ie, $B_q=1$). Fig.\ \ref{fig:efficiency}(d) reveals that test accuracy is unaffected even with a large $B_q=32$ (\ie, a single video per mini-batch), while training time and GPU memory usage drop by a significant 55\% and 42\%, approaching the upper bound by our analysis.

\smallskip \noindent \textbf{Video-centric \vs.\ query-centric inference.} We now ask to what extent the re-use of video representations enabled by SnAG's late fusion design accelerates inference. Fig.\ \ref{fig:efficiency}(b) presents the percentage of time spent by our video-centric inference scheme relative to query-centric inference, where the full model is evaluated on each and every video-query pair. Consistent with earlier findings on video-centric training, longer videos with more queries result in more saving in inference time. On the challenging MAD dataset, this alone brings a major reduction (65\%) in test time.

\smallskip \noindent \textbf{Inference speed.} With window filtering~\cite{hou2022cone}, inference of SnAG on the $val$ split of MAD takes 12 minutes on a single P100 GPU. This is 33\% faster than CONE~\cite{hou2022cone} ($\sim$18 mins) and 63\% faster than SOONet~\cite{pan2023scanning} ($\sim$32 mins).\footnote{SOONet reports an inference time of 8.2 mins on a single A100 GPU.}

% conclusion
\section{Conclusion}

In this paper, we presented a simple and accurate model towards scaling up video grounding. We provided extensive analysis on the complexity of video grounding models from the perspective of cross-modal fusion, and argued that late fusion is key to scalability in video duration and query size. Our model, SnAG, is a simple instantiation of late fusion with minimalist design. It achieves superior accuracy and efficiency on a spectrum of challenging benchmarks and compares favorably to strong baselines including those dedicated to efficient inference on long videos. We hope our analysis and model design can shed light on video grounding and more broadly scalable video understanding.

\smallskip
{\small \noindent{\bf Acknowledgement}:
This work was partially supported by NIH grant R01OH012313, by grants from McPherson Eye Research Institute and VCGRE at UW Madison, and by the Army Research Lab under contract number W911NF-2020221.}

{
%\small
\bibliographystyle{ieee_fullname}
\bibliography{egbib}
}

\clearpage
\section*{Supplementary Material}

\setcounter{figure}{0}
\setcounter{table}{0}
\setcounter{section}{0}
\setcounter{equation}{0}
\renewcommand{\theequation}{\Alph{equation}}
\renewcommand{\thefigure}{\Alph{figure}}
\renewcommand{\thesection}{\Alph{section}}
\renewcommand{\thetable}{\Alph{table}}

In the supplementary material, we (1) elaborate on our cost analysis in the main paper (Section~\ref{sec:appendix_cost}); (2) compare SnAG to latest methods with more advanced learning paradigms (Section~\ref{sec:appendix_comparisons}); (3) analyze the effect of input snippet length in training on model performance (Section~\ref{sec:appendix_ablation}); (4) conduct error analysis of our results in comparison to baseline results (Section~\ref{sec:appendix_error_analysis}); (5) describe additional implementation details (Section~\ref{sec:appendix_details}); and (6) discuss the limitations and future directions of our work (Section~\ref{sec:appendix_discussion}). We hope that this document will complement our main paper. 

For sections, figures, tables, and equations, we use numbers (\eg, Sec.\ 1) to refer to the main paper and capital letters (\eg, Sec.\ A) to refer to this supplement. 

\section{Cost Analysis}\label{sec:appendix_cost}

In Section~\ref{sec:method_inference} of the main paper, we have shown that late fusion, under the assumption of $\mathcal{C}(\mathcal{F}_L) \ll \mathcal{C}(g)+\mathcal{C}(h)$ and $\mathcal{C}(g) \ll \mathcal{C}(h)$, allows a maximum of $N$ times speedup in theory at inference time. Here $N$ is the number of text queries per video, $\mathcal{C}(g)$ and $\mathcal{C}(h)$ denotes the respective cost of video and text encoding, and $\mathcal{C}(\mathcal{F}_L)$ combines the cost of cross-modal fusion and moment decoding.

In practice, for any realistic instantiation of $\mathcal{F}_L$, like our SnAG, the cost of cross-modal fusion and moment decoding $\mathcal{C}(\mathcal{F}_L)$ can not be ignored. This partially breaks our assumption of $\mathcal{C}(\mathcal{F}_L) \ll \mathcal{C}(g)+\mathcal{C}(h)$. To account for non-trivial cost $\mathcal{C}(\mathcal{F}_L)$, we denote $\mathcal{C}(\mathcal{F}_L)=\alpha(\mathcal{C}(g)+\mathcal{C}(h))$, where the positive coefficient $\alpha$ reflects the cost of $\mathcal{F}_L$ relative to the total cost of both encoders $g$ and $h$. For late fusion, $\alpha$ is typically in the range of $(0, 1)$. Following Section~\ref{sec:method_inference}, we have
\begin{align}
\small
    \mathbf{R}_{\text{inf}} &= \frac{\mathcal{C}(\text{inference with early fusion})}{\mathcal{C}(\text{inference with late fusion})} \\
    &= \frac{M N \ \mathcal{C}(\mathcal{F}_{L}) + M N \ \mathcal{C}(g) + M N \ \mathcal{C}(h)}{M N\  \mathcal{C}(\mathcal{F}_{L}) + M\ \mathcal{C}(g) + N\ \mathcal{C}(h)} \\
    &= \frac{M N (1+\alpha)}{M N \alpha + M \mathbf{R}_{g/(g+h)} + N (1-\mathbf{R}_{g/(g+h)})} \\
    & \approx \frac{1+\alpha}{\frac{1}{N}+\alpha} \to
    \begin{cases}
        N, & \alpha\to 0 \\
		1+\frac{1}{\alpha}, & N\to \infty
    \end{cases}\label{eq:approx},
\end{align}
where the approximation in Equation~\ref{eq:approx} is a consequence of $\mathbf{R}_{g/(g+h)} = \mathcal{C}(g) / (\mathcal{C}(g) + \mathcal{C}(h)) \xrightarrow{\mathcal{C}(g) \gg \mathcal{C}(h)} 1$.

Substituting $N$ with $B_q$ (the number of queries drawn for each training snippet), we can derive similar efficiency result for our video-centric training scheme (Section~\ref{sec:method_training}):
\begin{align}
    \mathbf{R}_{\text{train}} \approx \frac{1+\alpha}{\frac{1}{B_q}+\alpha} \to
    \begin{cases}
        B_q, & \alpha\to 0 \\
		1+\frac{1}{\alpha}, & B_q\to \infty
    \end{cases}.
\end{align}

For long-form video grounding (\ie, large $N$ and $B_q$), our extended analysis indicates that late fusion remains highly efficient when $\alpha$ is reasonably small. For example, a late-fusion model with $\alpha \to 1$ can in theory bring a considerable 2$\times$ cost reduction in both training and inference relative to early fusion. This is precisely the case with SnAG, as Figure~\ref{fig:efficiency}(a) of the main paper shows that SnAG's encoders account for $\sim$50\% of the total computation.

Importantly, our empirical findings confirms this estimate. Figure~\ref{fig:efficiency}(b, d) of the main paper shows that video-centric training and inference of SnAG on MAD is $\sim$50\% more efficient in terms of runtime and GPU memory. On TACoS and Ego4D, the saving in training is less prominent as we found that smaller $B_q$ leads to slightly better results, yet the substantial reduction in inference time echoes our cost analysis and demonstrates the efficiency of SnAG.

\begin{table*}
\begin{subtable}[t]{\textwidth}
\begin{tabular}{lcccccccccccc}
\toprule[1.5pt]
\multirow{2}{*}{Model} & \multicolumn{3}{c}{R@1} & \multicolumn{3}{c}{R@5} & \multicolumn{3}{c}{R@10} & \multicolumn{3}{c}{R@50} \\ \cmidrule(r){2-4} \cmidrule(r){5-7} \cmidrule(r){5-7} \cmidrule(r){8-10} \cmidrule(r){11-13}
& 0.1  & 0.3  & 0.5  & 0.1   & 0.3   & 0.5  & 0.1   & 0.3   & 0.5   & 0.1   & 0.3   & 0.5 \\
\midrule[0.5pt]
$\dagger$ CLIP~\cite{radford2021clip} & 9.30 & 4.65 & 2.16 & 18.96 & 13.06 & 7.40 & 24.30 & 17.73 & 11.09  & 39.79 & 32.23 & 23.21 \\
$\dagger$ VLG-Net~\cite{soldan2021vlgnet} & 5.60 & 4.28 & 2.48 & 16.07 & 13.14 & 8.78 & 23.64 & 19.86 & 13.72  & 45.35 & 39.77 & 30.22 \\
$\dagger$ Moment-DETR~\cite{lei2021momentdetr} & 5.07 & 3.82 & 2.39 & 16.30 & 12.60 & 7.90 & 24.79 & 19.43 & 12.06 & 50.06 & 40.52 & 24.87 \\ 
\textbf{SnAG}  & \textbf{10.28} & \textbf{8.46} & \textbf{5.55} & \textbf{24.42} & \textbf{20.60} & \textbf{13.75} & \textbf{32.23} & \textbf{27.50} & \textbf{19.00} & \textbf{52.28} & \textbf{46.68} & \textbf{35.24} \\
\bottomrule[1.5pt]
\end{tabular}
\centering
\caption{\footnotesize Results on MAD~\cite{soldan2022mad}.}
\end{subtable}
\hspace{\fill}

\smallskip

\begin{subtable}[t]{\textwidth}
\begin{tabular}{lcc}
\toprule[1.5pt]
Model & mean R@1 & mean R@5 \\
\midrule[0.5pt]
 $\dagger$ Moment-DETR~\cite{lei2021momentdetr} & 7.28 & 22.14 \\
 $\dagger$ VSL-Net~\cite{zhang2020vslnet} & 11.90 & 24.06 \\
 \textbf{SnAG} (EgoVLP) & \textbf{15.59} & \textbf{38.40} \\
\bottomrule[1.5pt]
\end{tabular}
\centering
\caption{\footnotesize Results on Ego4D-NLQ~\cite{grauman2022ego4d}.}
\end{subtable}

\bigskip

\caption{\textbf{Comparison with Multimodal Guidance~\cite{fang2023tcsf}.} SnAG does not take audio as input, yet outperforms existing models with Multimodal Guidance ($\dagger$) using audio. Mean R@\{1,5\} are computed over tIoU=\{0.1,0.3,0.5\}. For SnAG, we report the better results between video-centric and query-centric training.}
\label{supp:table:mg}
\vspace{-1em}
\end{table*}

\begin{table}
\begin{subtable}[t]{\columnwidth}
\begin{tabular}{lcccc} 
\toprule[1.5pt]
\multirow{2}{*}{Model} & \multicolumn{2}{c}{R@1} & \multicolumn{2}{c}{R@5} \\
\cmidrule(r){2-3} \cmidrule(r){4-5}
 & 0.3 & 0.5 & 0.3 & 0.5 \\ 
\midrule[0.5pt]
 TCSF & 49.82 & 38.53 & 68.60 & 59.89 \\
 \textbf{SnAG} (C3D) & \textbf{56.44} & \textbf{44.86} & \textbf{81.15} & \textbf{70.66} \\
\bottomrule[1.5pt]
\end{tabular}
\centering
\caption{\footnotesize Results on TACoS~\cite{regneri2013tacos}.}
\end{subtable}
\hspace{\fill}

\smallskip

\begin{subtable}[t]{\columnwidth}
\begin{tabular}{lcccc} 
\toprule[1.5pt]
\multirow{2}{*}{Model} & \multicolumn{2}{c}{R@1} & \multicolumn{2}{c}{R@5} \\
\cmidrule(r){2-3} \cmidrule(r){4-5}
 & 0.5 & 0.7 & 0.5 & 0.7 \\ 
\midrule[0.5pt]
 TCSF & 53.85 & 37.20 & 90.86 & 58.95 \\
 \textbf{SnAG} (I3D) & \textbf{65.13} & \textbf{46.26} & \textbf{92.80} & \textbf{71.94} \\
\bottomrule[1.5pt]
\end{tabular}
\centering
\caption{\footnotesize Results on Charades-STA~\cite{sigurdsson2016charades}.}
\end{subtable}

\smallskip

\begin{subtable}[t]{\columnwidth}
\begin{tabular}{lcccc} 
\toprule[1.5pt]
\multirow{2}{*}{Model} & \multicolumn{2}{c}{R@1} & \multicolumn{2}{c}{R@5} \\
\cmidrule(r){2-3} \cmidrule(r){4-5}
 & 0.3 & 0.5 & 0.3 & 0.5 \\ 
\midrule[0.5pt]
 TCSF & \textbf{66.87} & 48.38 & 88.75 & 80.24 \\
 \textbf{SnAG} (C3D) & 63.58 & \textbf{48.55} & \textbf{89.55} & \textbf{81.71} \\
\bottomrule[1.5pt]
\end{tabular}
\centering
\caption{\footnotesize Results on ActivityNet-Captions~\cite{krishna2017activitynet}.}
\end{subtable}

\bigskip

\caption{\textbf{Comparison with TCSF~\cite{fang2023tcsf}.} SnAG outperforms TCSF despite using frozen video features. For SnAG, we report the better results between video-centric and query-centric training.}
\label{supp:table:tcsf}
\vspace{-1em}
\end{table}

\begin{table}
\begin{subtable}[t]{\columnwidth}
\begin{tabular}{lccc} 
\toprule[1.5pt]
\multirow{2}{*}{Model} & \multicolumn{3}{c}{R@1} \\
\cmidrule(r){2-4}
 & 0.3 & 0.5 & avg \\ 
\midrule[0.5pt]
 UniVTG & 7.28 & 3.95 & 5.62 \\
 UniVTG w/ PT & 11.74 & 7.54 & 9.64 \\
 \textbf{SnAG} (EgoVLP) & \textbf{15.87} & \textbf{11.26} & \textbf{13.57} \\
\bottomrule[1.5pt]
\end{tabular}
\centering
\caption{\footnotesize Results on Ego4D-NLQ~\cite{grauman2022ego4d}.}
\end{subtable}
\hspace{\fill}

\smallskip

\begin{subtable}[t]{\columnwidth}
\begin{tabular}{lccc} 
\toprule[1.5pt] 
\multirow{2}{*}{Model} & \multicolumn{3}{c}{R@1} \\
\cmidrule(r){2-4}
 & 0.3 & 0.5 & avg \\ 
\midrule[0.5pt]
 UniVTG & 51.44 & 34.97 & 43.21 \\
 UniVTG w/ PT & 56.11 & 43.44 & 49.78 \\
 \textbf{SnAG} (C3D) & \textbf{56.44} & \textbf{44.86} & \textbf{50.65} \\
\bottomrule[1.5pt]
\end{tabular}
\centering
\caption{\footnotesize Results on TACoS~\cite{regneri2013tacos}.}
\end{subtable}

\smallskip

\begin{subtable}[t]{\columnwidth}
\begin{tabular}{lccc} 
\toprule[1.5pt]   
\multirow{2}{*}{Model} & \multicolumn{3}{c}{R@1} \\
\cmidrule(r){2-4}
 & 0.5 & 0.7 & avg \\ 
\midrule[0.5pt]
 UniVTG & 58.01 & 35.65 & 46.83 \\
 UniVTG w/ PT & 60.19 & 38.55 & 49.27 \\
 \textbf{SnAG} (I3D) & \textbf{65.13} & \textbf{46.26} & \textbf{55.70} \\
\bottomrule[1.5pt]
\end{tabular}
\centering
\caption{\footnotesize Results on Charades-STA~\cite{sigurdsson2016charades}.}
\end{subtable}

\bigskip

\caption{\textbf{Comparison with UniVTG~\cite{lin2023univtg}.} SnAG consistently outperforms UniVTG despite being simpler and learning from individual datasets without multi-task pre-training. w/ PT means fine-tuning after pre-training. For SnAG, we report the better results between video-centric and query-centric training.}
\label{supp:table:univtg}
\vspace{-1em}
\end{table}

\section{Additional Comparisons}\label{sec:appendix_comparisons}

We further compare SnAG to latest approaches that (1) consume additional input modalities (\eg, audio), (2) are trained end to end on raw videos (as opposed to using frozen video backbones); or (3) perform large-scale, multi-task pre-training. We note that these are not fair comparisons and put SnAG at a disadvantage. This comparative analysis, however, further highlights the strength of SnAG.

\smallskip
\noindent \textbf{Comparison with Barrios \etal.~\cite{barrios2023localizing}.} Barrios \etal.~\cite{barrios2023localizing} has demonstrated a method for long-form video grounding via Multimodal Guidance. The guidance model takes audio in addition to video and text queries as input, and produces a score to condition an existing grounding model.

Table~\ref{supp:table:mg} compares the performance of existing models with Multimodal Guidance and SnAG. While the audio input provides additional cues to facilitate accurate moment localization, it does not fully close the gap between the existing models and SnAG. SnAG is 3.07 absolute percentage points higher in R@1, tIoU=0.5 on MAD, and 1.67 points higher in mean R@1 on Ego4D. We hypothesize that Multimodal Guidance will also boost the performance of SnAG.

\smallskip
\noindent \textbf{Comparison with TCSF~\cite{fang2023tcsf}.}
TCSF takes a compressed video and obtains features from keyframes without explicitly decoding the video. Importantly, the feature extractor is jointly trained with the grounding model. Joint training has significant benefit as it allows video features to adapt to the grounding task through end-to-end learning.

Table~\ref{supp:table:tcsf} compares the performance of TCSF and SnAG. Despite the extra capacity to learn more powerful features, TCSF lags behind SnAG by a significant margin. Consider the R@1, tIoU=0.5 scores --- SnAG outperforms TCSF by 6.33 absolute percentage points on TACoS, and 11.28 absolute percentage points on Charades. The two methods are on par on ActivityNet (48.38 \vs.\ 48.55).

\smallskip
\noindent \textbf{Comparison with UniVTG~\cite{lin2023univtg}.} UniVTG pools large-scale datasets and annotations from multiple video understanding tasks and learns a unified grounding model through multi-task pre-training. The pre-trained model supports fine-tuning and zero-shot inference on the training tasks, including temporal sentence grounding.

Table~\ref{supp:table:univtg} compares the performance of UniVTG and SnAG. While UniVTG is capable of learning from diverse data and supervision signals, its performance lags behind SnAG, despite using stronger SlowFast~\cite{feichtenhofer2019slowfast} and CLIP~\cite{radford2021clip} features. We again consider the R@1, tIoU=0.5 scores --- SnAG achieves 11.26\% on Ego4D, 44.85\% on TACoS, and 64.62\% on Charades, outperforming UniVTG by 3.72, 1.41 and 4.43 absolute percentage points, respectively.

\begin{figure}[!t]
\centering
\includegraphics[width=1.0\linewidth]{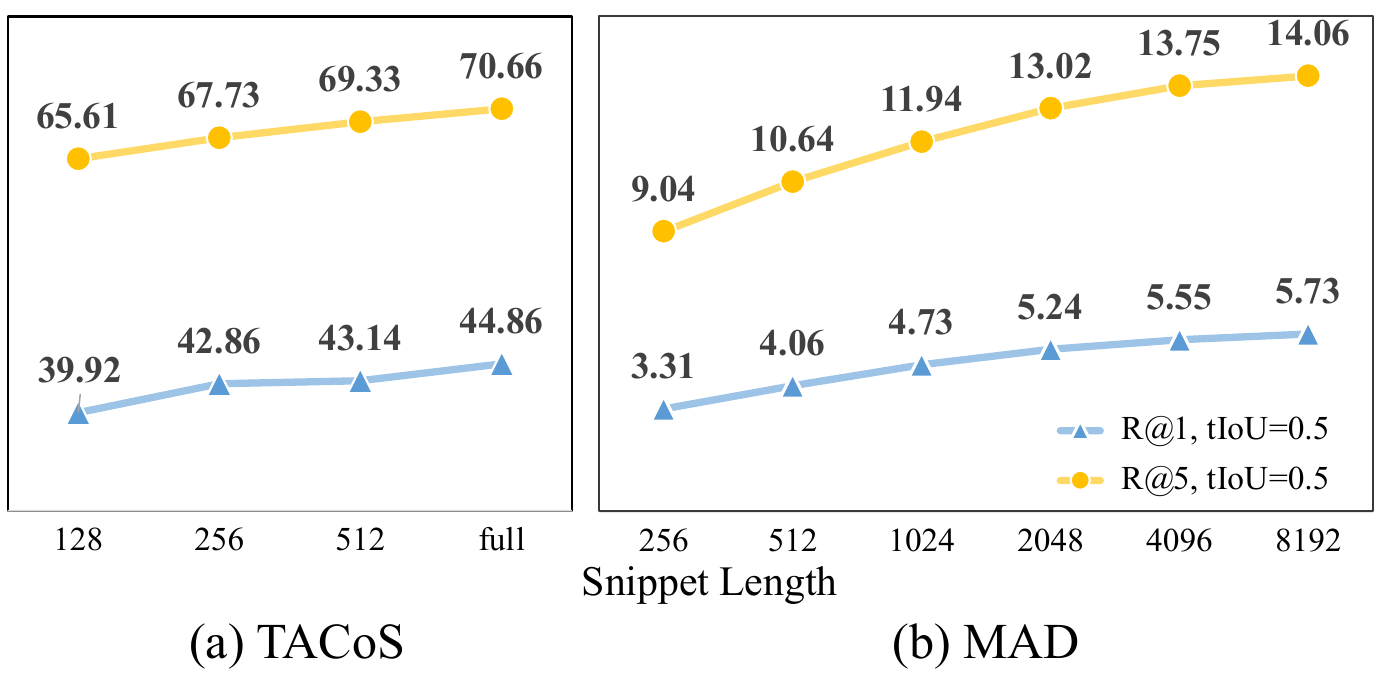}
\vspace{-1em}
\caption{\textbf{Ablations on snippet length in training.} On both TACoS (a) and MAD (b), SnAG delivers stronger test accuracy when trained on snippets of increased length.}
\label{fig:snippet_len}
\vspace{-1em}
\end{figure}

\section{Effect of Snippet Length}\label{sec:appendix_ablation}
We conduct ablation studies on MAD and TACoS to understand the impact of snippet length in training.

\smallskip
\noindent \textbf{Experiment setting.} On TACoS, we emulate feature sub-sampling in previous methods~\cite{soldan2021vlgnet,zhang20202dtan,liu2022mgslnet} by resizing input video features to a uniform length of 128, 256 and 512, and compare to training using full-sized features (capped at a length of 2304). On MAD, we are able to vary snippet length from 256 (1 minute) to 8192 (1 hour) thanks to SnAG's scalable model design and training scheme. We report R@1 and R@5 at tIoU=0.5 for both datasets.

\smallskip
\noindent \textbf{Results.}  Figure~\ref{fig:snippet_len} presents our results. Training on longer snippets constantly leads to higher test accuracy for both datasets. On MAD, R@1, tIoU=0.5 grows by a significant 73\% (from 3.31\% to 5.73\%) even though all moments are short (4.1 seconds on average) and can be fully covered by the shortest snippet length of 256. We conjecture that negative moments that are distant in time can provide strong learning signal to models like SnAG, which is capable of processing long-form video inputs.

\begin{figure*}[!t]
\centering
\includegraphics[width=0.95\linewidth]{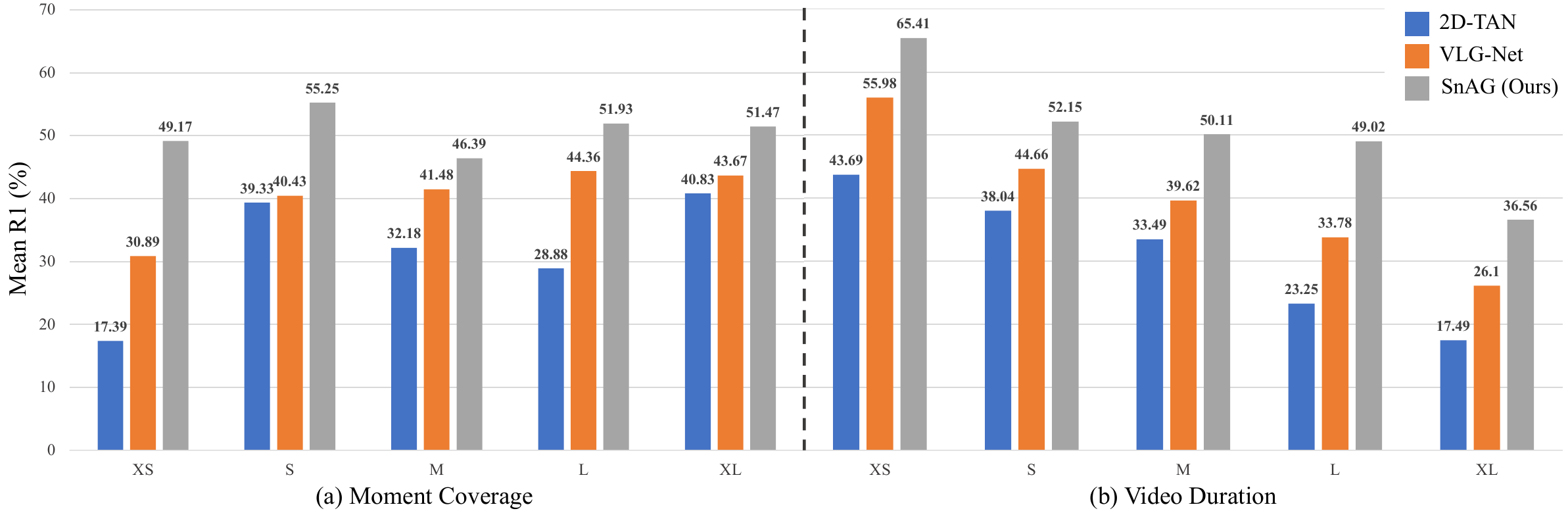}
\vspace{0.5em}
\caption{\textbf{Error analysis on TACoS.} Our model consistently outperforms VLG-Net~\cite{soldan2021vlgnet} and 2D-TAN~\cite{zhang20202dtan} on moments with low (XS) to high (XL) coverage and videos with short (XS) to long (XL) duration. It is especially strong on low-coverage moments (a) and long videos (b) thanks to its scalability with respect to video duration.}
\label{fig:error}
\vspace{-0.5em}
\end{figure*}

\section{Error Analysis}\label{sec:appendix_error_analysis}

To better understand our results, we present an error analysis and compare our results with those from strong baselines of VLG-Net~\cite{soldan2021vlgnet} and 2D-TAN~\cite{zhang20202dtan}. This analysis is conducted on the test split of TACoS.

\smallskip
\noindent \textbf{Experiment setting.} In our first experiment, we divide the ground-truth moments into five bins according to their coverage (\ie, moment length relative to full video length): Extra Small (XS: (0, 0.02]), Small (S: (0.02, 0.04]), Medium (M: (0.04, 0.08]), Large (L: (0.08, 0.16]) and Extra Large (XL: (0.16, 1]). In our second experiment, we divide the videos into five bins based on their duration in seconds: Extra Small (XS: (0, 125]), Small (S: (125, 250]), Medium (M: (250, 500]), Large (L: (500, 750]), Extra Large (XL: (750, $\infty$)). We report mean R@1 averaged over the tIoU thresholds of 0.3 and 0.5 for each bin.

\smallskip
\noindent \textbf{Results.} Figure~\ref{fig:error} presents our results. While our model consistently outperforms the baselines by a significant margin across all conditions, the relative percentage of improvement is especially large on low-coverage moments (XS and S) and long videos (L and XL). Note that both VLG-Net and 2D-TAN have a complexity of $\mathcal{O}(T^2)$ and thus need to sub-sample input video features to a small length (256 for VLG-Net and 128 for 2D-TAN) to reduce their computational cost. Intuitively, feature sub-sampling drops detailed information about event dynamics and makes precise localization of short moments more challenging. This becomes especially problematic for long videos as the drop rate can be extremely large (\eg, 10$\times$ for XL videos). Our results thus suggest that the ability to scale up to long videos is critical for enhancing localization accuracy.

% \usepackage{multirow}
\begin{table*}[]
\resizebox{\linewidth}{17mm}{
\begin{tabular}{l||cccccccc|ccc|cccc}
\toprule[1.5pt]
Dataset                       & \begin{tabular}[c]{@{}c@{}}Video\\FPS\end{tabular} & \begin{tabular}[c]{@{}c@{}}Video\\  Feature\end{tabular} & \begin{tabular}[c]{@{}c@{}}Clip\\  Size\end{tabular} & Stride              & \begin{tabular}[c]{@{}c@{}}Snippet\\  Length\end{tabular} & \begin{tabular}[c]{@{}c@{}}Video\\  Dim\end{tabular} & \begin{tabular}[c]{@{}c@{}}Win\\  Size\end{tabular} & \begin{tabular}[c]{@{}c@{}}Video\\  Layers\end{tabular} & \begin{tabular}[c]{@{}c@{}}Text\\  Feature\end{tabular} & \begin{tabular}[c]{@{}c@{}}Text\\  Dim\end{tabular} & \begin{tabular}[c]{@{}c@{}}Text\\  Layers\end{tabular} & \begin{tabular}[c]{@{}c@{}}Batch\\  Size\end{tabular} & $B_q$ & lr                        & Epochs              \\ \hline
MAD                           & 5                                                    & CLIP                                                     & 1                                                    & 1                   & 4096                                                      & 512                                                  & 17                                                  & 9                                                       & CLIP                                                    & 512                                                 & 1                                                      & 32                                                    & 8  & 1e-4                  & 8                   \\
\multirow{2}{*}{Ego4D-NLQ}    & \multirow{2}{*}{30}                                  & SlowFast                                                 & \multirow{2}{*}{32}                                  & \multirow{2}{*}{16} & \multirow{2}{*}{2304}                                     & 512                                                  & \multirow{2}{*}{19}                                 & \multirow{2}{*}{8}                                      & BERT                                                    & 128                                                 & \multirow{2}{*}{1}                                     & 2                                                     & 2  & \multirow{2}{*}{1e-4} & 10                  \\ 
                              &                                                      & EgoVLP                                                   &                                                      &                     &                                                           & 384                                                  &                                                     &                                                         & EgoVLP                                                  & 768                                                 &                                                        & 2                                                     & 2  &                           & 7                   \\
TACoS                         & 29.4                                                 & C3D                                                      & 16                                                   & 16                  & 2304                                                      & 128                                                  & 19                                                  & 8                                                       & GloVe                                                   & 128                                                 & 5                                                      & 16                                                    & 4  & 1e-3                  & 15                  \\
\multirow{2}{*}{Charades-STA} & \multirow{2}{*}{24}                                  & C3D                                                      & \multirow{2}{*}{16}                                  & \multirow{2}{*}{4}  & \multirow{2}{*}{256}                                      & \multirow{2}{*}{256}                                 & \multirow{2}{*}{5}                                  & \multirow{2}{*}{7}                                      & \multirow{2}{*}{GloVe}                                  & \multirow{2}{*}{128}                                & \multirow{2}{*}{5}                                     & 16                                                    & 4  & \multirow{2}{*}{1e-3} & \multirow{2}{*}{10} \\
                              &                                                      & I3D                                                      &                                                      &                     &                                                           &                                                      &                                                     &                                                         &                                                         &                                                     &                                                        & 16                                                    & 4  &                           &                     \\
ANet-Captions                 & varying                                              & C3D                                                      & 16                                                   & 8                   & 256                                                       & 128                                                  & 5                                                   & 7                                                       & GloVe                                                   & 128                                                 & 5                                                      & 16                                                    & 2  & 1e-3                  & 15     \\
\bottomrule[1.5pt]
\end{tabular}
}\vspace{0.5em}
\caption{\textbf{Implementation Details.} We list key parameters for video encoding (left), text encoding (middle), and model training (right).}
\label{table:implementation}
\vspace{-1em}
\end{table*}

\section{Implementation Details}\label{sec:appendix_details}

We now present the implementation details of SnAG. Table~\ref{table:implementation} summarizes key hyper-parameters we used in our experiments for feature extraction, training and inference.

\smallskip
\noindent \textbf{Video features.} For TACoS~\cite{regneri2013tacos}, Charades-STA~\cite{sigurdsson2016charades} and ActivityNet-Captions~\cite{krishna2017activitynet}, we extract clip-level video features using C3D network pre-trained on Sports-1M~\cite{tran2015c3d}, following prior approaches (\eg,~\cite{zhang20202dtan,soldan2021vlgnet,liu2022mgslnet}). 

We decode TACoS videos at a frame rate of 29.4 frames per second (FPS) and compute the 4096-dimensional, post-ReLU activations of \emph{fc6} layer as features using non-overlapping clips of 16 frames. For Charades-STA, we use the RGB frames provided on the official website. We use clips of 16 frames with a stride of 4 frames (\ie, 75\% overlap), and compute the 4096-dimensional, post-ReLU activations of \emph{fc7} layer as features. We use official C3D features for ActivityNet. For fair comparison with baselines, we additionally extract two-stream I3D features for Charades-STA from I3D network pre-trained on Kinetics~\cite{carreira2017i3d} using clips of 16 frames with a stride of 4 frames. For Ego4D-NLQ~\cite{grauman2022ego4d}, we use official SlowFast features~\cite{feichtenhofer2019slowfast}as well as EgoVLP features~\cite{lin2022egovlp}, both extracted from 30 FPS videos using clips of 32 frames with a stride of 16 frames (\ie, 50\% overlap). For MAD~\cite{soldan2022mad}, we use the official CLIP features~\cite{radford2021clip} extracted at 5 FPS.

\smallskip
\noindent \textbf{Textual features.} We use GloVe word embeddings~\cite{pennington2014glove} with 6B vocabulary for TACoS, Charades-STA and ActivityNet-Captions. For Ego4D-NLQ, we use last-layer token embeddings ($\mathrm{CLS}$ token excluded) from BERT~\cite{devlin2018bert} or EgoVLP text encoder together with SlowFast or EgoVLP video features. For MAD, we use official CLIP token embeddings ($\mathrm{SOT}$ and $\mathrm{EOT}$ tokens excluded) from the last layer of text encoder.

\smallskip
\noindent \textbf{Network architecture.} For the video encoder, we apply a linear projection followed by two 1D convolutional layers with kernel size 3 on the input features prior to building the multi-scale video representation. The embedding dimension, attention window size and number of scales in the multi-scale Transformer network differ by dataset (see Table~\ref{table:implementation} for a summary). For the text encoder, we apply a single Transformer layer for MAD and Ego4D-NLQ, and 5 layers for the other datasets. We use 4 attention heads for all Transformer layers in the model. The classification and regression heads each contains two 1D convolutional layers with kernel size 3.

\smallskip
\noindent \textbf{Training details.} We randomly sample video snippets with a maximum sequence length of 4096, 2304 and 2304 for MAD, Ego4D-NLQ and TACoS, respectively. We resize all videos to a uniform length of 256 for Charades-STA and ActivityNet. We use the AdamW optimizer~\cite{loshchilov2019decoupled} with the default beta values of (0.9, 0.999) and a weight decay of 0.05. Mini-batch size (including $B_q$ for video-centric training), learning rate and number of training epochs can be found in Table~\ref{table:implementation}. We maintain an exponential moving average (EMA) of model parameters over the entire course of training with a momentum of 0.999 and use the EMA version of the trained model for inference.

\smallskip
\noindent \textbf{Inference details.} Following our video-centric inference protocol, we first feed full videos into our model to compute the shared video representation, and then input text queries and fuse them with the video representation. Moments are then decoded from the fused representation and further merged using SoftNMS~\cite{bodla2017softnms}.

\section{Discussion and Limitation}\label{sec:appendix_discussion}

The design of SnAG is one of many possible instantiations of our analysis. As long as late fusion and video-centric training is employed, alternative designs can be considered, including two-stage proposal-based methods or DETR-alike models. We anticipate that exploration of this design space may point to an interesting future avenue. 

Similar to prior methods, SnAG's performance heavily depends on the quality of pre-extracted video and sentence features. These features are from networks pre-trained on a related yet different task (\eg, action recognition or image-text matching) and may not be optimal for video grounding. Training of SnAG further requires moment and sentence annotations that are expensive and difficult to acquire. End-to-end training~\cite{cheng2022stochastic} of SnAG with less human annotation~\cite{mithun2019weakly} is thus a promising future direction.

\end{document}